\documentclass[10pt,twocolumn,letterpaper]{article}

\usepackage[pagenumbers]{cvpr} 

\usepackage[accsupp]{axessibility}  

\usepackage{graphicx}
\usepackage{amsmath}
\usepackage{amssymb}
\usepackage{booktabs}
\usepackage{soul}
\usepackage{verbatim}
\usepackage{multirow}
\usepackage{boldline}
\usepackage{enumitem}
\usepackage{bm}

\usepackage[pagebackref,breaklinks,colorlinks]{hyperref}

\usepackage[capitalize]{cleveref}
\crefname{section}{Sec.}{Secs.}
\Crefname{section}{Section}{Sections}
\Crefname{table}{Table}{Tables}
\crefname{table}{Tab.}{Tabs.}

\def\cvprPaperID{16} 
\def\confName{CVPR}
\def\confYear{2022}

\begin{document}

\title{K-Lane: Lidar Lane Dataset and Benchmark for Urban Roads and Highways}

\author{Dong-Hee Paek$^{1}$\thanks{co-first authors} \quad\quad Seung-Hyun Kong$^{1}$\footnotemark[1] \thanks{corresponding author} \quad\quad Kevin Tirta Wijaya$^{2}$\\
$^{1}$CCS Graduate School of Mobility\quad$^{2}$Robotics Program\\
Korea Advanced Institute of Science and Technology, Daejeon, Korea\\ 
{\tt\small \{donghee.paek, skong, kevin.tirta\}@kaist.ac.kr}
}
\maketitle

\begin{abstract}
Lane detection is a critical function for autonomous driving. 
With the recent development of deep learning and the publication of camera lane datasets and benchmarks, camera lane detection networks (CLDNs) have been remarkably developed.
Unfortunately, CLDNs rely on camera images which are often distorted near the vanishing line and prone to poor lighting condition.
This is in contrast with Lidar lane detection networks (LLDNs), which can directly extract the lane lines on the bird’s eye view (BEV) for motion planning and operate robustly under various lighting conditions.
However, LLDNs have not been actively studied, mostly due to the absence of large public lidar lane datasets. 
In this paper, we introduce KAIST-Lane (K-Lane), the world’s first and the largest public urban road and highway lane dataset for Lidar. 
K-Lane has more than 15K frames and contains annotations of up to six lanes under various road and traffic conditions, e.g., occluded roads of multiple occlusion levels, roads at day and night times, merging (converging and diverging) and curved lanes.
We also provide baseline networks we term Lidar lane detection networks utilizing global feature correlator (LLDN-GFC).
LLDN-GFC exploits the spatial characteristics of lane lines on the point cloud, which are sparse, thin, and stretched along the entire ground plane of the point cloud.
From experimental results, LLDN-GFC achieves the state-of-the-art performance with an F1-score of 82.1\%, on the K-Lane. 
Moreover, LLDN-GFC shows strong performance under various lighting conditions, which is unlike CLDNs, and also robust even in the case of severe occlusions, unlike LLDNs using the conventional CNN.
The K-Lane, LLDN-GFC training code, pre-trained models, and complete development kits including evaluation, visualization and annotation tools are available at \href{https://github.com/kaist-avelab/k-lane}{\color{black}{https://github.com/kaist-avelab/k-lane}}.
\end{abstract}

\section{Introduction}

\begin{figure*}[htb!]
\centering
\includegraphics[width=1.9\columnwidth]{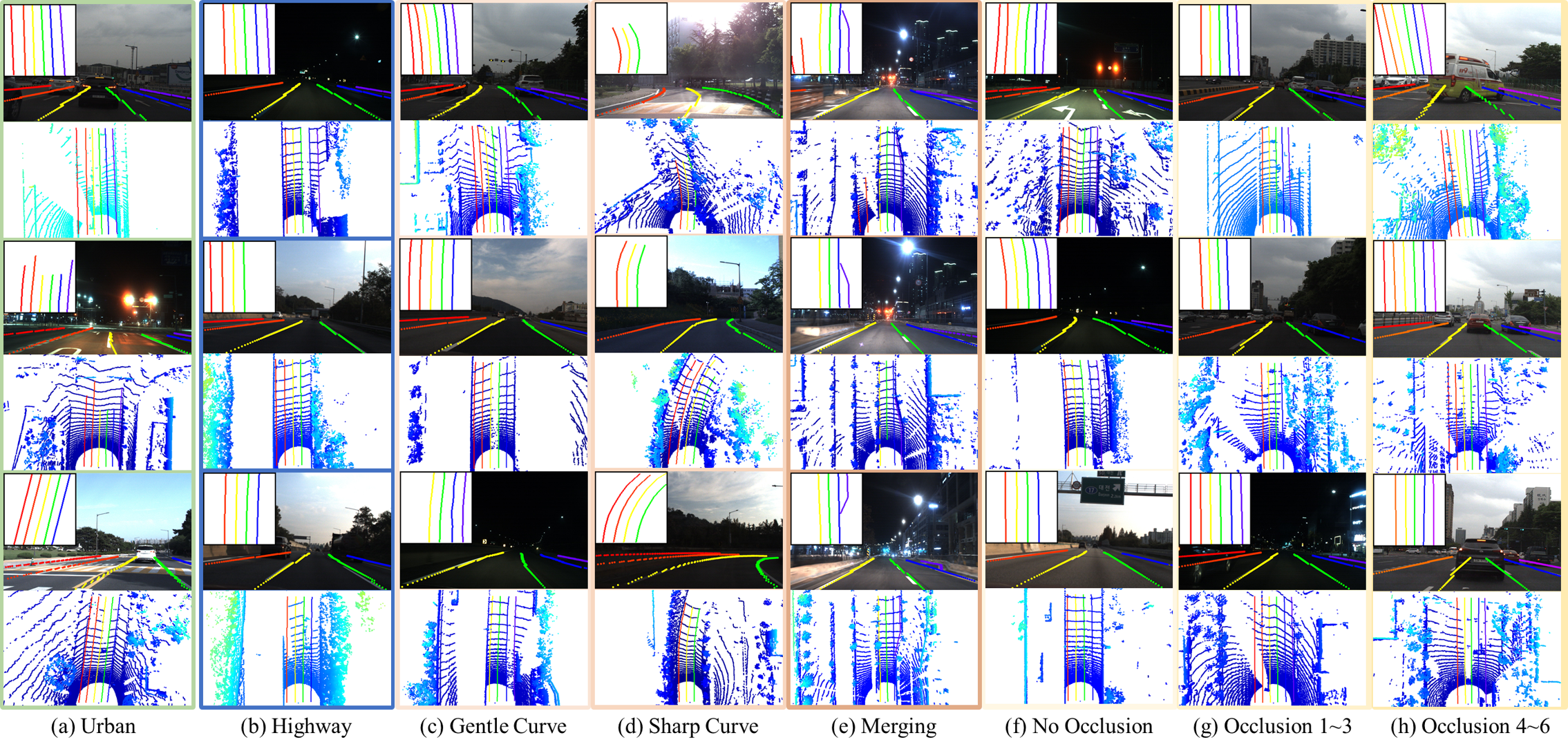}
\caption{Examples of frames under various conditions for K-Lane, where each column shows one of the conditions: Each column consists of a total of three rows. Each row shows an upper plot for the projection of true BEV labels into the front view image with true BEV labels in the upper left corner and a lower plot for the lane labels on top of the BEV point cloud, respectively.}
\label{klane}
\end{figure*}

Autonomous driving depends on a number of critical functions that are realized with state-of-the-art (SOTA) technologies.
Among them, lane detection function is to detect the accurate location and curvature of the ego lane and neighboring lanes, and provide necessary input to the path planning function. 
Therefore, the lane detection function should be robust to various conditions (e.g., night, day times) and challenging situations (e.g., lane line occlusion). 
However, the conventional lane detection techniques based on image processing are vulnerable to situations when lane lines are partially missing or occluded, because the techniques rely on heuristic methods such as denoising, edge detection, and line fitting with detected edges \cite{heu_cam_1,heu_cam_2,heu_cam_3}.

Recently, lane detection \cite{scnn,condlane,laneatt} has been largely improved due to the deep learning. 
When a large dataset with accurate label is available for training, deep learning networks can produce high-quality predictions that are almost indistinguishable to the ground truths.
This is the case for camera lane detection networks (CLDNs) \cite{laneatt,condlane}, which show superior performances compared to the conventional (i.e., heuristic) lane detection techniques when given abundant training samples from public datasets such as CULane \cite{scnn} and TuSimple \cite{tusimple}.

However, CLDNs still have a few inherent problems. First, cameras suffer from poor lighting conditions, such as low or harsh lights \cite{scnn}. 
Second, it is often necessary to project front camera images into 2-dimensional (2D) bird’s eye view (BEV) for motion planning, which often causes lane line distortions \cite{lane_fusion}. 
For example, BEV-projection of the detected lanes near the vanishing line of a front camera image \cite{bev_camera} may result in inaccurate and distorted lane lines so that the motion planning should be limited to a shorter distance.

On the other hand, Lidar has multiple advantages over the camera in the lane detection; lane detection from a Lidar point cloud does not suffer from the distortion in the BEV-projection and is not affected by lighting conditions. 
However, there have been a little studies introduced in the literature, mostly because there are not enough public dataset and benchmark for Lidar lane detection network (LLDN).

In this paper, we introduce KAIST-Lane (K-Lane) dataset, the world’s first and the largest open Lidar lane dataset for Lidar lane detection in urban roads and highways. 
We also provide an easy-to-use development kits (devkits) for the training, evaluation, dataset development, and visualization.  K-Lane has more than 15K annotated frames, and contains a maximum of six lanes under various road and traffic conditions, such as roads at night and day times, merging (converging and diverging) and curved lanes as shown in Fig. \ref{klane}. 
Each annotation consists of the lane lines segmentation label, driving condition, lane shape, and occlusion level.
As such, the performance of developed LLDNs in different challenging conditions can be evaluated easily, e.g., when driving in the night time or with significant measurement loss due to high occlusions.
The segmentation label is accurately annotated with one pixel width on the BEV image, which translates to a 4cm $\times$ 4cm area in the real world.
The label consists of a class id, which represent the relative position of the lane line to the ego-lane.
This enables the LLDNs to be trained directly to infer the location of ego-lane, which is crucial for the motion planning.
Moreover, as shown in Fig. \ref{klane}, front camera images have been elaborately calibrated with Lidar point clouds, enabling intuitive visualization and may pave the way for further lane detection studies using multi-modal sensors (e.g., sensor fusion). 

To demonstrate the viability of developing LLDNs with K-Lane, we propose a baseline model, Lidar lane detection network utilizing global feature correlator (LLDN-GFC), which fully exploits the spatial characteristics of the lane lines in point clouds.
This is in contrast to most of the CNN-based LLDNs introduced in the literature \cite{deeplidar_egolane,lane_fusion}, which are mostly a modification of the CNN-based CLDNs developed for camera images. 
We observe that the CNN-based LLDNs are not suitable for detecting lane lines in Lidar point cloud.
For example, lane lines on the front view image have decreasing thickness with the distance from the ego vehicle and are heading to the same vanishing point (on a straight road), whereas lane lines in a BEV image have a constant thickness and stretch long in parallel over the entire BEV image. 
These spatial characteristics of the lane lines in Lidar point cloud are not appropriately exploited by the CNN-based lane detection networks, in contrast to our proposed LLDN-GFC. 
The proposed LLDN-GFC can be implemented with Transformer \cite{vit} and Mixer \cite{mixer} blocks to perform an effective global feature correlation for lane lines.
In the experimental results, we show that the proposed baseline achieves a superior performance than LLDNs using the conventional CNN. The contribution of this paper can be summarized as;

\begin{table*}[htb!]
{
\small
\centering
\begin{tabular}{c|c|c|c|c|c|c|c}
\hlineB{2}
Dataset  & Year & Sensor & Num. Lanes & Class Info. & Num. Frames & Public & Road Type \\ \hlineB{2}
TuSimple \cite{tusimple} & 2017 & Camera & 5 & No & 6408 & Yes & Highway \\ \hline
CULane \cite{scnn} & 2018 & Camera & 4 & Yes & 133235 & Yes & Urban \& Highway \\ \hline
DeepLane \cite{lane_fusion} & 2019 & Lidar, Camera & n/a & No & 55168 & No & Urban \& Highway \\ \hline
RoadNet \cite{deeplidar_egolane} & 2020 & Lidar & n/a & Yes & 5200 & No & Highway \\ \hline
K-Lane (ours) & 2022 & Lidar, Camera & 6 & Yes & 15382 & Yes & Urban \& Highway \\ \hlineB{2}
\end{tabular}
\caption{Comparison of previous lane datasets and ours (n/a denotes not available).}
\label{tab:dataset}
}
\end{table*}

\begin{itemize}[noitemsep]
    \item K-Lane: we introduce the world first and the largest (15382 frames) public Lidar lane dataset for urban roads and highways under various conditions and scenarios.
    \item We also provide a complete devkits for training, evaluation, annotation, and visualization.
    \item We show that lane lines in the Lidar point cloud have a special characteristics not found in traditional RGB images, and provide appropriate baseline network we term LLDN-GFC, which significantly outperforms LLDNs using the conventional CNN in the F1-score.
\end{itemize}

This paper is organized as follows. Section 2 introduces prior studies related to this paper and the topic of this paper, Section 3 introduces K-Lane dataset, and the proposed baseline, LLDN-GFC. Section 4 shows the experiment setup and results. We draw our conclusion in Section 5, and introduce more information for both dataset and baseline, such as detailed network structure of LLDN-GFC in the Appendix.


\begin{figure*}[htb!]
\centering
\includegraphics[width=1.65\columnwidth]{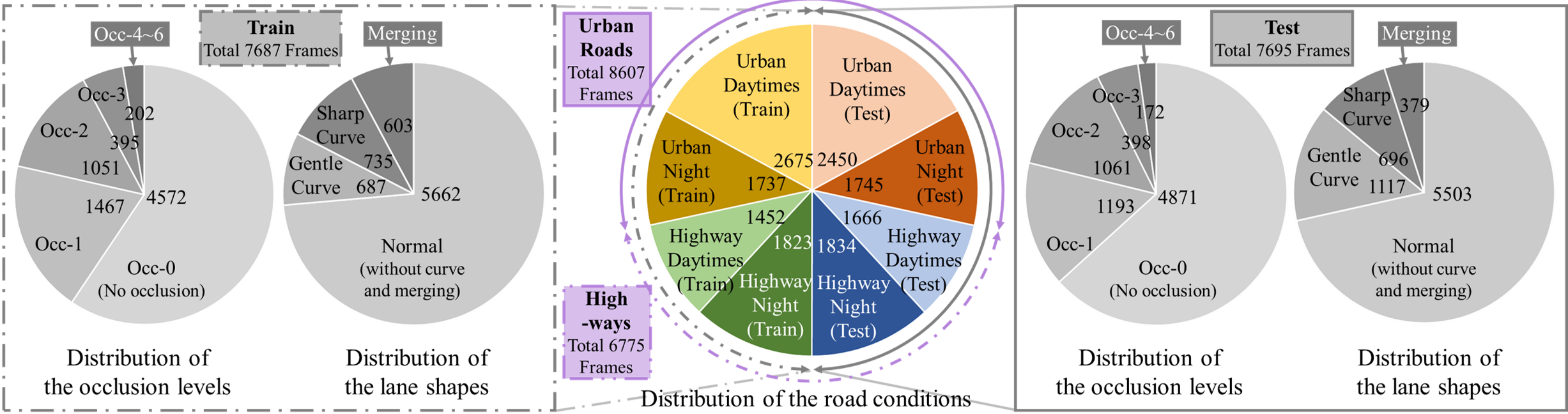}
\caption{Data distribution (in number of data frames) of the K-Lane. The pie-chart in the middle shows data distribution in road types (i.e., urban roads or highways), time (i.e., daytimes or night), and usage (i.e., training or test). The four pie-charts on the left (for training) and right (for test) shows data distribution with respect to the six levels of lane occlusion (from zero occluded lanes to six occluded lanes) and that with respect to the lane shape (gentle curve, sharp curve and merging).}
\label{klane_dist}
\end{figure*}

\section{Related Works}

\noindent\textbf{Lane Detection Datasets and Benchmarks.}
Lane detection with data-driven approaches such as deep learning has seen tremendous advancements in recent years.
One key enabler towards such advancements is the availability of large public lane dataset, as shown in Table \ref{tab:dataset}. 
TuSimple \cite{tusimple} is one of the earliest publicly available camera-based lane dataset.
It has 6,408 number of frames collected in the highway during the day.
The dataset is further divided into 3,626 frames for training, 358 frames for validation, and 2,782 frames for testing.
CULane \cite{scnn} introduces a more diverse and challenging camera-based lane dataset, with 133,235 number of frames divided into 88,880 frames for training, 9,675 frames for validation, and 34,680 frames for testing.
CULane provides diverse driving conditions, both in urban and highway environments, in the day and night, and with various road structures.
Comparing to the vibrant field of camera-based lane detection, Lidar lane detection dataset has not been explored as much.
One of the earliest Lidar lane dataset is DeepLane \cite{lane_fusion}, which contains 55,168 frames of Lidar and camera data collected in both urban and highway environments.
Another dataset, RoadNet \cite{deeplidar_egolane}, consists of 5,200 frames of Lidar data collected only in the highway environments.
Unfortunatelly, both datasets are not public, as such not many derivative works on Lidar lane detection have been conducted.
In contrast, our proposed dataset, K-Lane, contains 15,382 frames of Lidar and camera data, collected in both urban and highway environements.
As we make K-Lane public, we pave the way for a new research direction in lane detection approaches using Lidar.

\noindent\textbf{Lane Detection Networks for Camera.}
As labeled camera dataset \cite{scnn} for various roads become available, there have been a significant advancement in the CLDNs. 
Compared with the early rule-based techniques \cite{heu_cam_2,heu_cam_3}, CLDNs are more adaptive to various road environments.
In these techniques, lanes are predicted based on local features extracted by CNN \cite{resnet}, and the performance is improved with lane detection heads that exploit the features of lane lines.
For example, Qin et al. \cite{ufast} proposes a row-wise detection-based network that divides the entire image into grids, and recognize lanes from each row of grids. 
Liu et al. \cite{condlane} proposes a two-stage lane detection network that combines the conditional convolution \cite{condconv} with the row-wise detection in the detection head and achieves SOTA performance in several datasets.
However, CLDNs have some inherent problems. 
In the CULane benchmark, most of the CLDNs show significant performance drop (about 20\%) for night time and dazzling light conditions from their daytime performance \cite{condlane,ufast}.

\noindent\textbf{Early Lane Detection Techniques for Lidar.}
In early studies, lane points are detected by thresholding the measured intensity (or reflectivity).
Lindner et al. \cite{heu_lidar_2} uses a fixed polar grid map to store point intensities and filter the lane candidates with thresholding along azimuth angles.
Hernandez et al. \cite{heu_lidar_3} introduces a clustering approach, where the filtered lane points are clustered using DBSCAN \cite{dbscan}.
However, these heuristic techniques rely on pre-defined thresholding parameters, and, therefore, it is not very adaptive to different environments.

\noindent\textbf{Lane Detection Networks for Lidar.}
Deep learning-based lane detection studies for Lidar have not been actively conducted due to the absence of large open datasets, and only some studies with their private Lidar datasets are introduced in the literature.
Bai et al. \cite{lane_fusion} proposes an LLDN that combines 2D BEV images developed with the Lidar point cloud and the front camera image for lane detection.
And Martinek et al. \cite{deeplidar_egolane} proposes a CNN-based LLDN that uses BEV images from point clouds to detect ego-lanes, and tests the network in an uncrowded highway.

\noindent\textbf{Self-Attention in Vision.}
Self-attention is a scheme to lead a neural network to pay more attention to the patches of the input image, between which there is high correlation score.
Convolutional block attention module (CBAM) \cite{cbam} introduces per-channel and per-space self-attention mechanisms by adding MLP (Multilayer Perceptron) and convolutional operations, respectively, to the traditional CNN-based feature extractor.
Since Transformer \cite{transformer} shows significant improvement in the self-attention mechanism by applying three independent MLPs for query, key, and value (i.e., Transformer block), it has been used actively for images and point cloud.
As an example, ViT (Vision Transformer) \cite{vit} greatly improves image classification performance using Transformer, where ViT divides input image is into unit patches and applies Transformer encoder to each patch for image classification.
However, ViT employs three independent MLPs for each attention mechanism, for which high computational cost and large model size are inevitable.
On the other hand, MLP-Mixer \cite{mixer} implements the attention mechanism with a simple MLP scheme (i.e., Mixer block), which results in a fast inference with a small model size and achieves comparable performance to ViT.


\section{K-Lane and LLDN-GFC}
In this section, we introduce K-Lane dataset, benchmark, and the proposed baseline, LLDN-GFC.

\subsection{K-Lane}

K-Lane is the first large open LiDAR lane dataset that consists of Lidar point clouds and their corresponding RGB images for urban roads and highways under various conditions and scenarios as shown in Fig. \ref{klane}.

\noindent\textbf{Data Distribution.}
As shown in Fig. \ref{klane_dist}, there are a total of 15382 data frames, divided into 7687 frames for training and 7695 frames for testing.
Each set contains various road conditions and challenging scenarios including 
(a) different lighting conditions such as day and night times, 
(b) crowded traffic with lane occlusions by other vehicles, and 
(c) merging (converging, diverging) and curved lanes, which are further classified into gentle curves and sharp curves.
Note that K-Lane has maximum six lanes and occlusions are divided into six levels representing 0, 1, 2, 3, and 4$\sim$6 occluded lanes.
The benchmark kit provides evaluation tools for calculating the metrics per each condition, and given conditions are annotated for each frame under a clear criterion, which are described in Appendix A.

\begin{figure*}[htb!]
\centering
\includegraphics[width=1.65\columnwidth]{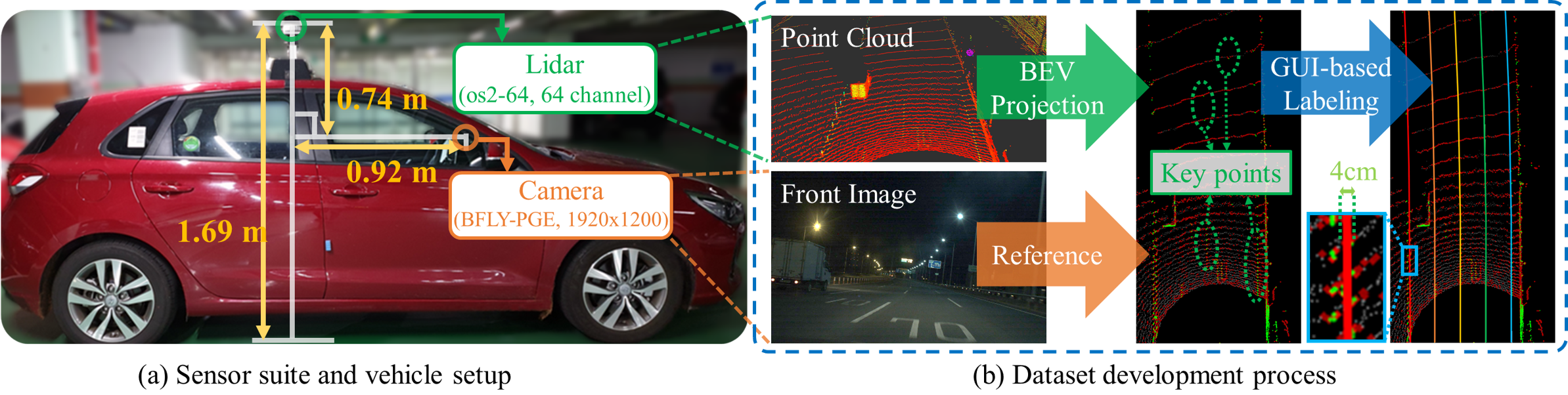}
\caption{Sensor suite, vehicle setup, and dataset development process of K-Lane.}
\label{sensor_suite_and_label}
\end{figure*}

\noindent\textbf{Sensor Suite.}
The K-Lane is collected using Ouster OS2-64 Lidar sensor \cite{ouster} that has 64 channels with a maximum range of 240m, placed on the roof of the vehicle, and a front camera that has $1920\times1200$ resolution, as shown in Fig. \ref{sensor_suite_and_label}-a. 
Front camera images have been carefully calibrated with Lidar point clouds, which makes it easy to visualize and may enable further lane detection studies with multi-modal sensors.

\noindent\textbf{Dataset Development.}
The ground truth labels are produced by projecting the Lidar point cloud into BEV, thresholding the intensity measurements to extract keypoints (i.e., candidates of lane lines), and drawing one-pixel-wide line for each lane as shown in Fig \ref{sensor_suite_and_label}-b.
As such, high resolution and accurate labels are produced, which is critical for deep learning-based methods.

\noindent\textbf{Metrics.}
To standardize the evaluation of the network being developed, we choose to use the F1-score metric for both confidence and classification, which evaluates per-pixel presence of lane and per-pixel correct classification of the lane, respectively.
The F1 metric represents a harmonic mean between precision and recall, and can be expressed as

\begin{equation}
\small
    \textnormal{F1} = \frac{2}{\frac{1}{\textnormal{Precision}} + \frac{1}{\textnormal{Recall}}} = \frac{\textnormal{TP}}{\textnormal{TP + 0.5(FP + FN)}},
    \label{f1_score}
\end{equation}

\noindent where TP, FP, and FN are the numbers of true positives, false positives, and false negatives, of the output of the detection head, respectively. 
Since the width of a lane line in the label is only one pixel-wide, we allow up to one pixel deviation between the prediction and the label.
This is comparable to the evaluation metric used in the CULane \cite{scnn} dataset, where a lane line label is 30-pixels wide and a true positive is counted when the prediction and the ground truth have at least an IoU of 0.5.

To formally describe the evaluation metric, let $\bm{x}^{conf} \in \mathbb{R}^{M\times N}$ be the confidence map label with $M$ number of rows and $N$ number of columns, with $x^{conf}_{m,n} \in \{0, 1\}$.
Furthermore, let $\hat{\bm{x}}^{conf} \in \mathbb{R}^{M \times N}$ be the confidence map prediction with $M$ number of rows and $N$ number of columns, with $x^{conf}_{m,n} \in [0, 1]$.
Additionally, we define a grid neighborhood centered at the grid $x_{m,n}$ as a set of grids $\{x_{i,j} | i = \{m-1, m, m+1\}, j = \{n-1, n, n+1\}\}$.

Suppose that a thresholding operation is applied to the confidence map prediction such that

\begin{equation}
\small
  \hat{x}^{conf, thr}_{m,n} = \left\{ 
  \begin{array}{ c l }
    1 & \quad \hat{x}^{conf}_{m,n} > \sigma^{conf} \\
    0                 & \quad \textrm{otherwise}
  \end{array},
\right.
\end{equation}

\noindent where the $\sigma_{conf}$ is the confidence threshold for a prediction to be considered as a lane point.
In our evaluation metric, a true positive occurs if for a positive prediction (pixel value equals to 1) at $\hat{x}^{conf, thr}_{m,n}$, there exists at least one positive label at the grid neighborhood centered around $x^{conf}_{m,n}$. 
Conversely, if there is no positive label at the grid neighborhood, the prediction counts as a false positive.
A false negative occurs if for a positive label at $x^{conf}_{m,n}$, there is no positive prediction at the grid neighborhood centered around $\hat{x}^{conf, thr}_{m,n}$.

For classification predictions, we transform the classification map label into a one-hot-encoding label.
In addition, we also transform the classification prediction map into a one-hot-encoding prediction where the class with the highest probability is assigned a value of 1 and the rest are set as 0.
The true positives, false positives, and false negatives of the classification predictions can be calculated with the previously mentioned procedure, accumulated for every possible classes.

The F1-score on classification is an evaluation of networks based on both lane line localization and lane class prediction. 
Therefore, F1-score on classification is a strict evaluation metric and, as a result, performance degradation can be found for all models compared to the confidence prediction, as shown in Table \ref{tab:conf-overall}.

\noindent\textbf{Complete Devkits.}
Additionally, we provide a comprehensive devkits of K-Lane that includes training, evaluation, dataset development, and visualization.
In particular, data development tools, such as labeling and annotation tools, are provided through the Graphic User Interface (GUI) for easy-to-use.
This enables the research community to readily increase the dataset regardless of the Lidar sensor models, and thus to activate areas of LLDN with diverse datasets and benchmarks, as well as CLDN.
Appendix A presents a full description of all of the specifics.

\noindent\textbf{Summary.}
In summary, compared to the conventional lane detection datasets, K-Lane has multiple advantages; 
(1) K-Lane is collected in urban roads and highways under various conditions and scenarios as stated above, while TuSimple \cite{tusimple} and RoadNet \cite{deeplidar_egolane} include only highway, 
(2) K-Lane distinguishes lane classes and labels with precise lane location (pixel level), whereas TuSimple \cite{tusimple} and DeepLane \cite{lane_fusion} have labels without distinction between lane classes
(3) K-Lane has larger number of labeled lanes (e.g., maximum 6 lanes), while TuSimple \cite{tusimple}, and CULane \cite{scnn} have only up to 5 and 4 lanes, respectively, and 
(4) Above all, among Lidar lane datasets, K-Lane is the only publicly available dataset, which allows more studies on Lidar-based lane detection to be conducted. In addition, the well-calibrated camera images may also be used in future works for multimodal lane detection.

\begin{figure}[b!]
\centering
\includegraphics[width=1.0\columnwidth]{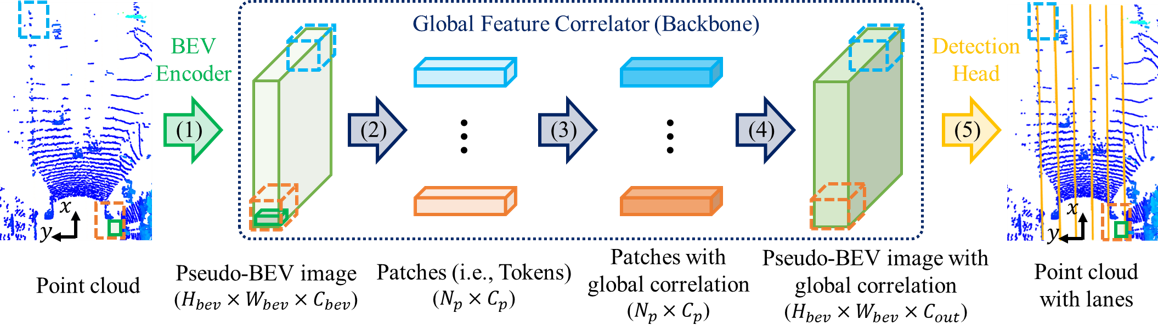}
\caption{Overall structure of the LLDN-GFC. There are five functions: (1), (2), (3), (4), and (5) indicate the BEV encoder, reshape \& per-patch linear transform, Transformer or Mixer block, reshape \& shared MLP, and detection head, respectively. $H_{bev}$, $W_{bev}$, $C_{bev}$, $C_{out}$, $N_p^2$, and $C_p$, are the height, width, num. of channel of the pseudo-BEV image, num. of channel of the output pseudo-BEV image, the num. of total patches, and the num. of channel per patch in the global correlation, respectively.}
\label{lldn_structure}
\end{figure}


\subsection{LLDN-GFC}

In this section, we focus on the overall structure and necessity of our baseline for LLDNs while the details, such as the exact neural network structure, functions (i.e., (1)$\sim$(5) in Fig. \ref{lldn_structure}), and mathematical expression for loss, are described in Appendix B. As shown in Fig. \ref{lldn_structure}, the proposed baseline consists of a BEV encoder, a GFC as backbone, and a lane detection head, which are introduced in the following subsections.

\noindent\textbf{BEV Encoder.} The BEV encoder projects a 3D point cloud into a 2D pseudo-image and process it further to produce a 2D BEV feature map. 
We provide two variants of BEV encoder for the LLDN-GFC, namely point projector and pillar encoder.

The primary BEV encoder is the point projector \cite{joint_projector,complex_yolo} that projects point clouds into the xy-horizontal plane and produces a BEV feature map using CNN. 
In order to maintain both high-resolution lane information and real-time speed, we design a ResNet-based CNN to output a feature map that is $1$/$8^2$ of the pseudo-image input.

An alternative for low computational 2D BEV encoder is the pillar encoder based on Point Pillars that has relatively small network size \cite{pillars}. 
Pillar encoder has slightly lower performance but faster inference speed than the CNN-based point projector. 
Therefore, in this paper, pillar encoder is presented as an alternative for real-time applications.
Details are in Appendix B.

\begin{figure}[tb!]
\centering
\includegraphics[width=0.95\columnwidth]{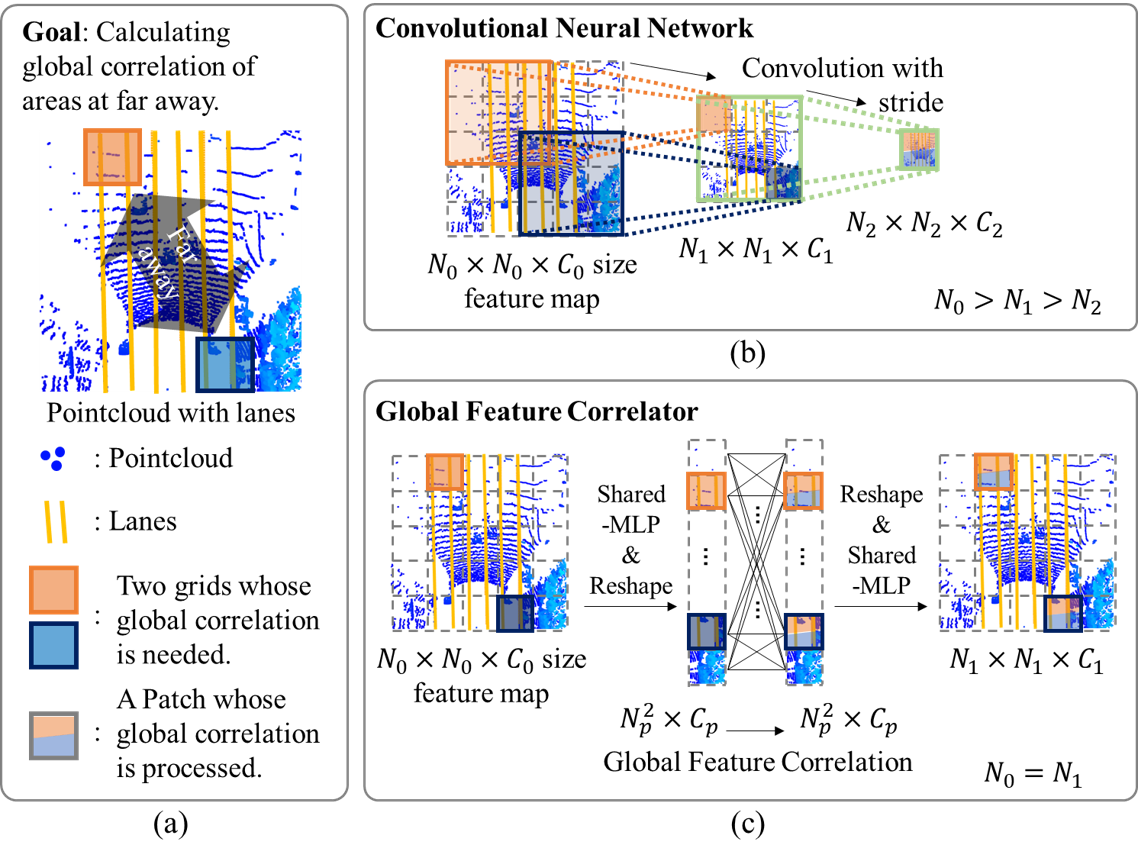}
\caption{A comparison of global feature correlation between CNN and the proposed GFC, where $N_0,N_1,N_2$ and $C_0,C_1,C_2$ represent the size and the number of channels of the feature maps at three layers in depth order. (a) an example of two separated grids to calculate the global correlation, (b) the two grids contained in a feature map developed by CNN, and (c) correlation of the two grids in a feature map developed by GFC.}
\label{cnn_vs_gfc}
\end{figure}

\noindent\textbf{GFC as Backbone.}
As shown in Fig. \ref{cnn_vs_gfc}-a, lane lines on the road are thin, stretched along the entire point cloud, and only occupy a small number of pixels (i.e., sparse).
Due to such thinness and sparsity, it is necessary to perform feature extraction in high resolution. 
In addition, the feature extractor should consider the correlation between distant grids within the BEV feature map.
As such, we design our proposed GFC to calculate global correlations of the features in high resolution by utilizing patch-wise self-attention networks.
We propose two variants of GFC: GFC-T (the main proposal based on Transformer blocks \cite{vit}) and GFC-M (the low computational alternative based on Mixer blocks \cite{mixer})

A major advantage of using patch-wise self attention networks is their capability to find correlations between distant grids (or patches) right from the early stages of backbone, as shown in Fig. \ref{cnn_vs_gfc}-c.
As such, the high-resolution information can be preserved (i.e., ${N_0}={N_1}$).
This is in contrast with CNN-based feature extractors, which may find correlations between distant grids after several layers of convolutions and down samplings, thus lowering the resolution of information (i.e., ${N_0}\gg{N_2}$), as shown in Fig. \ref{cnn_vs_gfc}-b.

Quantitatively, we observe that patch-wise self-attention networks have higher performance compared to their CNN-based counterparts \cite{deeplidar_egolane}.
In addition, we visualize the qualitative results of intermediate feature maps and attention scores in Fig. \ref{heatmap} and \ref{attention}, respectively.
Both quantitative and qualitative results further indicate the aptness of using patch-wise self-attention networks for Lidar lane detection even on a relatively small number of data (i.e. 7687 training frames).

\noindent\textbf{Detection Head and Loss Function.}
To design the detection head, we formulize the lane detection problem as a multi-class segmentation problem, where each pixel is assigned a class and a confidence score. 
The multi-class segmentation formulation enables the detection head to predict both lane classes and various lane shapes, which are important for motion planning where the ego vehicle need to plan inter-lane motions or recognize lane merging and separation. 
The LLDN-GFC detection head consists of two segmentation heads, each of which consists of a sequence of two-layer shared-MLP with a non-linear activation in-between.

As the number of lane samples are significantly smaller than the number of background samples on each frame, we incorporate the soft dice loss \cite{diceloss} for the confidence loss that inherently handles the imbalance problem.
For the classification head, we choose the grid-wise cross-entropy loss \cite{unet} that has been widely used for multi-class classification problems, leading the network to learn to maximize the probability of the correct lane class during training. 
The total loss function is the summation of both the soft-dice loss and the cross-entropy loss as expressed in Appendix B.

\begin{table*}[ht]
{
\centering
\small
\begin{tabular}{cc|c|c|c|c|c|c|c|c|c}
\hlineB{3}
\multicolumn{1}{c|}{Enc} & Backbone & Total & Day & Night & Urban & Highway & Shp Curve & No Occ & Occ4$\sim$6 & FPS \\ \hlineB{3}
\multicolumn{1}{c|}{\multirow{5}{*}{Proj-28}} & GFC-T3 & \textbf{82.1/81.1} & \textbf{82.2/81.4} & \textbf{82.0/80.7} & \textbf{81.7/80.6} & \textbf{82.5/81.7} & \textbf{78.0/76.7} & \textbf{82.9/81.9} & \textbf{75.9/75.5} & 11.6 \\ \cline{2-11} 
\multicolumn{1}{c|}{} & GFC-M3 & 79.7/78.8 & 79.9/79.0 & 79.6/78.4 & 78.9/77.7 & 80.8/80.0 & 74.6/72.6 & 80.4/79.4 & 72.5/71.7 & 13.3 \\ \cline{2-11} 
\multicolumn{1}{c|}{} & RNF-S13 & 73.2/70.5 & 72.6/70.1 & 74.0/71.0 & 73.1/70.4 & 73.3/70.6 & 70.5/68.1 & 74.9/72.3 & 63.5/59.0 & 13.1 \\ \cline{2-11} 
\multicolumn{1}{c|}{} & RNF-C13 & 78.0/75.3 & 77.6/75.1 & 78.5/75.5 & 77.7/74.8 & 78.3/76.01 & 76.0/73.1 & 79.6/77.0 & 69.3/65.3 & 13.0 \\ \cline{2-11} 
\multicolumn{1}{c|}{} & RNF-D23 & 72.1/68.8 & 71.3/68.3 & 73.0/69.4 & 71.9/68.7 & 72.3/69.0 & 69.6/66.5 & 74.0/70.7 & 61.9/57.6 & 12.7 \\ \hlineB{2}
\multicolumn{1}{c|}{\multirow{5}{*}{Pillars}} & GFC-T5 & 78.5/77.3 & 78.5/77.6 & 78.4/77.0 & 77.8/76.4 & 79.2/78.4 & 72.5/70.2 & 79.4/78.2 & 70.2/69.5 & 13.8 \\ \cline{2-11} 
\multicolumn{1}{c|}{} & GFC-M5 & 74.8/73.5 & 74.8/73.6 & 74.9/73.4 & 72.0/70.5 & 78.2/77.1 & 64.6/62.2 & 75.5/74.2 & 65.2/62.3 & \textbf{16.3} \\ \cline{2-11} 
\multicolumn{1}{c|}{} & RNF-S13 & 64.6/18.2 & 62.9/16.4 & 66.5/20.4 & 59.4/15.6 & 70.7/21.4 & 51.1/13.1 & 65.5/19.2 & 44.9/4.7 & 15.7 \\ \cline{2-11} 
\multicolumn{1}{c|}{} & RNF-C13 & 76.8/40.6 & 75.9/39.1 & 77.8/42.4 & 74.5/40.6 & 79.6/40.6 & 67.6/32.5 & 77.9/43.6 & 62.5/20.4 & 15.5 \\ \cline{2-11} 
\multicolumn{1}{c|}{} & RNF-D23 & 63.2/19.0 & 61.6/17.0 & 65.0/21.2 & 58.6/17.2 & 68.6/21.1 & 51.1/13.5 & 64.2/20.2 & 43.9/4.9 & 15.2 \\ \hlineB{2}
\multicolumn{2}{c|}{Heuristic} & \multicolumn{1}{c|}{26.4} & \multicolumn{1}{c|}{26.8} & \multicolumn{1}{c|}{26.0} & \multicolumn{1}{c|}{23.3} & \multicolumn{1}{c|}{29.6} & \multicolumn{1}{c|}{27.6} & \multicolumn{1}{c|}{28.1} & \multicolumn{1}{c|}{16.7} & \multicolumn{1}{c}{18.3} \\ \hlineB{3}
\end{tabular}
\caption{\label{tab:conf-overall}F1-score of confidence/classification for the proposed LLDN-GFC and various CNN-based LLDNs. 
Enc, Shp, Occ stands for BEV Encoder, sharp curve, and occlusion cases, respectively.
We show no occlusion and severe occlusion (4$\sim$6 lanes occluded) cases, while other occlusion levels are presented in Appendix C.
FPS stands for frame per second, which represents the overall computational cost (FLOPs, data efficiency, etc.) of the networks during inference, similar to throughput in \cite{mixer}.
Note that we only show F1-score of confidence for the heuristic technique.}
}
\end{table*}

\section{Experiments and Comparison}
In this section, we provide detailed performance comparisons between LLDN-GFC and conventional CNN-based LLDNs. 
In addition, we also discuss recent CLDNs for a general comparison to the LLDN-GFC performance.

\noindent\textbf{Implementation Details.}
We evaluate two variants of LLDN-GFC, Proj28-GFC-T3 and Pillars-GFC-M5, which we observe during experiments (i.e., ablations in Appendix C) to have the best accuracy and speed-accuracy tradeoff, respectively.
Proj28-GFC-T3 stands for LLDN-GFC with point projector encoder with 28 layers and GFC with three Transformer blocks.
Pillars-GFC-M5 stands for LLDN-GFC with pillar encoder and GFC with five Mixer blocks.

We use RTX3090 GPUs for training the networks on the K-Lane for 60 epochs using Adam optimizer \cite{adam} with a batch size 4 and a learning rate of 0.0002. 
All training and evaluations are implemented with PyTorch 1.7.1 \cite{pytorch} on an Ubuntu 18.04 machine.

\subsection{LLDN-GFC vs. CNN-based LLDN}

We consider three types of CNN-based backbone to be compared with the proposed GFC, namely: RNF-S13, RNF-D23, and RNF-C13, where (1) RNF represent ResNet \cite{resnet} with feature pyramid network (FPN) \cite{fpn}, (2) S13, D23, and C13 represent residual blocks implemented with strided convolution, dilated convolution, and CBAM \cite{cbam} of 13 or 23 layers, respectively.
The model capacities of the counterparts are also determined with experiments (i.e., ablations in Appendix C).
The FPN is applied to synthetically consider feature maps of different levels, and the dilated convolution increase the receptive field without loss of resolution, which is utilized in the existing LLDN \cite{deeplidar_egolane}.
The CBAM performs self-attention mechanism similar to the proposed LLDN-GFC, but applied with per-channel convolution operation, meaning that it does not perform global correlations for all patches as in LLDN-GFC.
For this reason, LLDN with RNF-C shows lower performance than the proposed LLDN-GFC, as shown in Table \ref{tab:conf-overall}.

\begin{figure}[b!]
\centering
\includegraphics[width=1.0\columnwidth]{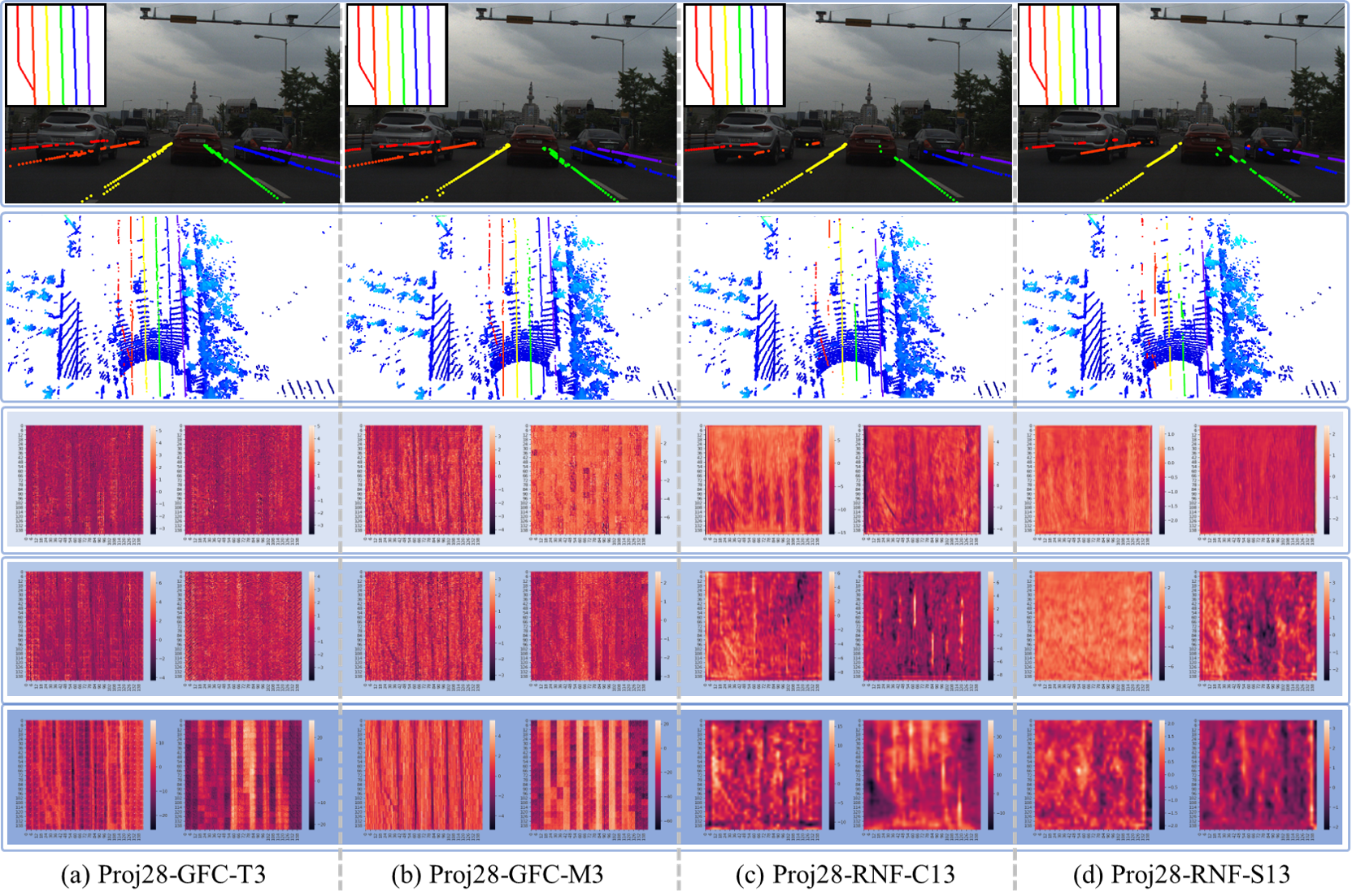}
\caption{Comparison of the lane detection performance between the LLDN-GFC and CNN-based LLDNs for occluded lanes condition. 
The four columns are inference results of (a) Proj28-GFC-T3, (b) Proj28-GFC-M3, (c) Proj28-RNF-C13, and (d) Proj28-RNF-S13. 
The first row shows the projection of inference results onto the front view image with labels in the upper left corner, while the second row shows the inference on the BEV point cloud. 
From the third to fifth row, we show the heatmaps sampled along the channels of the 1st, 2nd, and 3rd block output feature map of the backbones (e.g., 1st , 2nd, and 3rd Transformer blocks of GFC or residual blocks of RNF).
Heatmaps for various scenarios such as curved and merging lane lines are introduced in Appendix D.
}
\label{heatmap}
\end{figure}

As summarized in Table \ref{tab:conf-overall}, the proposed LLDN-GFC shows superior performance than the LLDNs with conventional CNN-based backbone of various depths. 
In particular, LLDN-GFC shows robust performance against sever occlusions, where four or more lanes are occluded.

Fig. \ref{heatmap} show qualitative assessment of the robustness of LLDN-GFC based on the visualization of intermediate feature maps.
We can observe on the heatmaps that both Proj28-GFC-T3 (a) and Proj28-GFC-M3 (b) clearly extract lanes with better resolution, especially on the deeper layers.
This is in contrast with CNN-based LLDN, shown in (c) and (d), where the lanes tend to blur.
In other words, the lane features extracted by Proj28-GFC-T3 and Proj28-GFC-M3 are more distinctive to the backgrounds compared to the lane features extracted by CNN-based LLDNs.
In addition, even in the presence of occlusions, Proj28-GFC-T3 and Proj28-GFC-M3 are capable of predicting the lane shapes through correlations with the non-occluded lanes, which are not observed in the CNN-based LLDNs.

\subsection{LLDN-GFC Attention Visualization}
In this subsection, we discuss the robustness of the proposed LLDN-GFC, Proj28-GFC-T3, against the occlusion scenario using the visualization of attention score.

\begin{figure}[b]
\centering
\includegraphics[width=0.99\columnwidth]{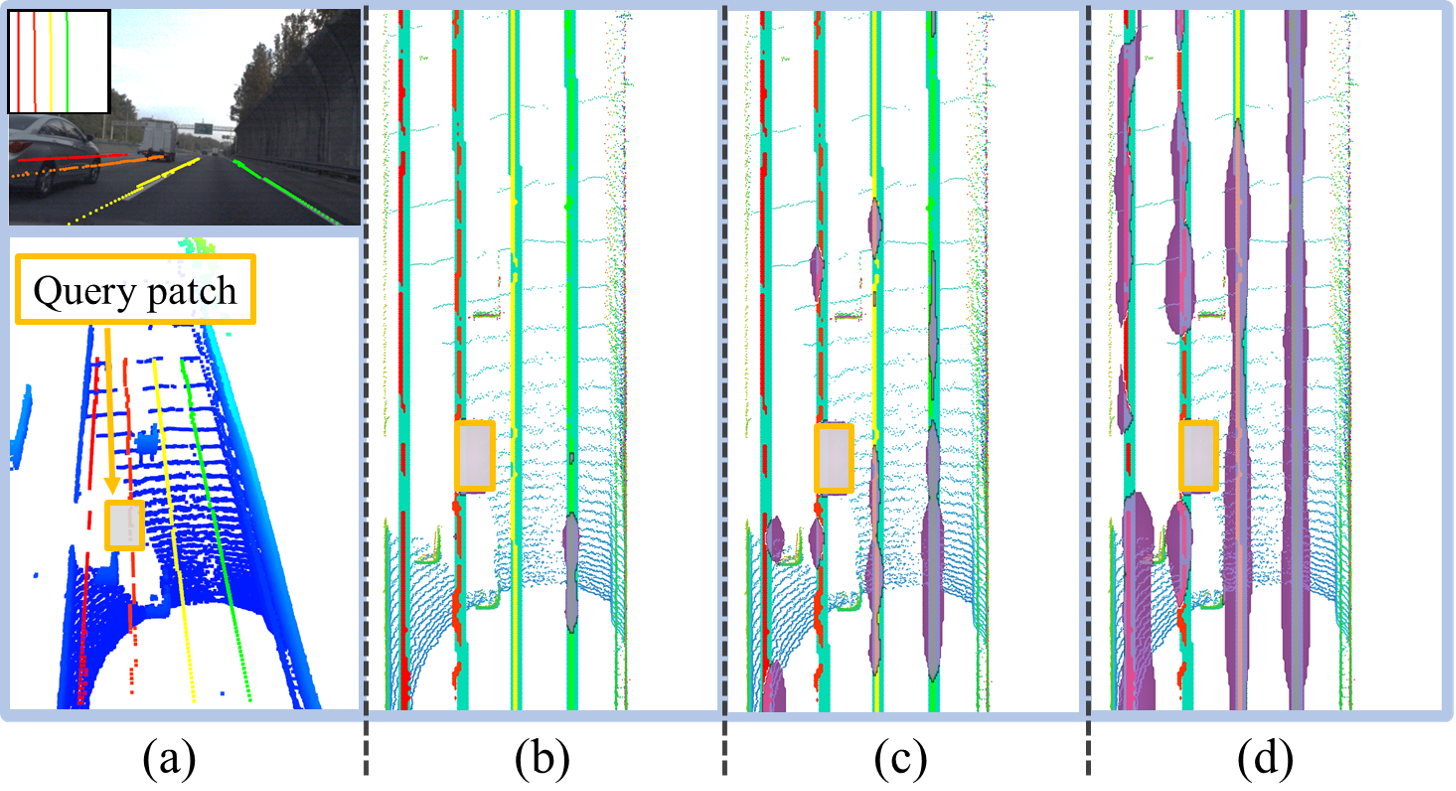}
\caption{Attention score visualization of Proj28-GFC-T3. 
(a-upper) shows the projection of inference results onto the front view image with labels in the upper left corner, while (a-lower) shows the inference result and the current query patch (yellow box) on top of the BEV point cloud.
(b) to (d) show the point cloud in BEV, lane inference results, query patch, labels in cyan, and attention score in purple for block 1, 2, and 3 of the GFC, respectively.}
\label{attention}
\end{figure}

The proposed GFC is based on the self-attention mechanism that utilizes correlations between data units to make the network give more attention to the meaningful region on the feature map.
As such, we can see the region that is considered as important by the GFC-T3 by visualizing the attention score prouced by each Transformer blocks, as shown in Fig. \ref{attention}.

From the visualization, we can see that the network give more attention to the regions which contain lane lines by attenuating the magnitude of non-lane-lines (irrelevant) features.
As the layers get deeper, the network expands its region of interests, indicated by the increasing area with high attention scores.
We observe that utilizing three blocks of transformer for GFC-T3 is sufficient to ensure the self-attention mechanism to cover the entire region of the point cloud which contains lane lines.
Additionally, note that for the query location (the yellow box in Fig \ref{attention}), the network produce high attention scores to regions in which lane lines are present, even if the query location is occluded.
Such phenomena indicates the robustness of LLDN-GFC to occlusions, where predictions are made by considering the entire point cloud such that occluded lane lines can still be estimated accurately.
This may not be possible for CNN-based LLDN, which features are recognized through local convolutions.

\subsection{LLDN-GFC vs. Camera Lane Detection}
Most state-of-the-art lane detection networks in the literature are for front-view camera images.
This means most CLDNs are trained to detect lane lines in the front-view map.
On the other hand, LLDN-GFC is trained to detect lane lines in the BEV map.
In addition, the environment in which the data are collected is different.
CULane is composed of data mostly for urban roads, while K-Lane consists of data for both urban roads and highways.
Since these Lidar and camera datasets do not use the same representations and are collected in different environments, we cannot compare the CLDNs simply using the reported performance in the literature. 

However, recent CLDNs show a significant performance drop for the night time data comparing to the daytime data. 
For example, CondLaneNet-Large \cite{condlane}, LaneATT-Large \cite{laneatt}, and CurveLane-NAS-L \cite{curvelane} show 18.67\%,  20.93\%, and 21.8\% drops between daytime and nighttime conditionss, respectively. 
In contrast, as shown in Table 1, the proposed LLDN-GFC shows almost no performance degradation (only 0.2\% difference). 
This is because the Lidar is robust to light conditions, which demonstrates that LLDN is a reliable function for autonomous driving.

\section{Conclusion}

In this work, we introduce K-Lane which, to the best of our knowledge, is the world's first publicly available dataset for Lidar lane detection.
K-Lane consists of over 15K high-quality annotated Lidar data in diverse and challenging driving conditions, along with well-callibrated front-view RGB images.
The driving conditions include various lighting (daytime and nighttime), lane occlusions (up to 6 occluded lane lines), and road structures (merging, diverging, curved lanes).
In addition, we provide the development kits for K-Lane including the annotation, visualization, training, and benchmarking tools.
We also introduce a baseline network for Lidar lane detection, which we term LLDN-GFC.
LLDN-GFC utilizes self-attention mechanisms to extract lane features via global correlation, and show superior performance compared to the conventional CNN-based LLDNs.
In addition, we show the importance of Lidar lane detection networks, where there is only little performance degradation in between detection in the daytime and detection in the nighttime, in contrast to camera-based lane detection networks.
As such, we expect this work to pave the way for further studies in the field of Lidar lane detection, and improve the safety aspects in autonomous driving.

\section*{Acknowledgment}

This work was supported by the National Research Foundation of Korea (NRF) grant funded by the Korea government (MSIT) (No. 2021R1A2C3008370)

\crefname{section}{Sec.}{Secs.}
\Crefname{section}{Section}{Sections}
\Crefname{table}{Table}{Tables}
\crefname{table}{Tab.}{Tabs.}

\newcounter{appdxTableint}
\newcounter{appdxFigureint}
\newcommand\tabcounterint{%
  \refstepcounter{appdxTableint}%
  \renewcommand{\thetable}{\arabic{appdxTableint}}%
}
\newcommand\figcounterint{%
  \refstepcounter{appdxFigureint}%
  \renewcommand{\thefigure}{\arabic{appdxFigureint}}%
}
\setcounter{appdxFigureint}{7}
\setcounter{appdxTableint}{2}

{\Large\noindent\textbf{Appendix}}
\vspace{2mm}

We provide a detailed description of the K-Lane dataset and the development kits (devkits), and detailed structure of proposed LLDN-GFC with its CNN-based counterparts, in Section A, and B, respectively.
In addition, Section C shows ablation study for the network hyper-parameters of the proposed LLDN-GFC (i.e., Proj28-GFC-T3), low computational alternative (i.e., Pillars-GFC-M5), and the counterparts.
Furthermore, Section D shows qualitative lane detection results for K-Lane, and visualization of both heatmap of features and the attention score.
Lastly, Section E shows the comparison between LLDN-GFC and heuristic lane detection methods.

\vspace{5mm}
{\large\noindent\textbf{A. K-Lane and Devkits}}
\vspace{2mm}

Section A contains technical details that may helps researchers in using the K-Lane datasets and devkits.

\vspace{2mm}
{\large\noindent\textbf{A.1. Details of K-Lane and Devkits}}
\vspace{2mm}

In this section, we present three additional details about the K-Lane: sequence distributions, compositions, and the criteria of driving conditions annotations of the dataset.

{\noindent\textbf{Sequence Distribution.}} K-Lane dataset consists of fifteen sequences that have different set of road conditions. The details of the sequences are shown in Table \ref{tab:sequence}. For the test data, we provide additional driving conditions annotations on each frame (i.e., curve, occlusion, merging, and number of lanes) with annotation tool shown in Section A.2.

{\noindent\textbf{Conditions Criteria.}} To evaluate the LLDN performance depending on data characteristics, we provide 13 different categories of driving conditions as shown in Table \ref{tab:a_4}.  Examples of each condition are shown Fig. \ref{klane}, and each frame can have two or more conditions, for example, day time and occlusion.

\begin{figure}[htb!]
{
\figcounterint
\centering
\includegraphics[width=1.0\columnwidth]{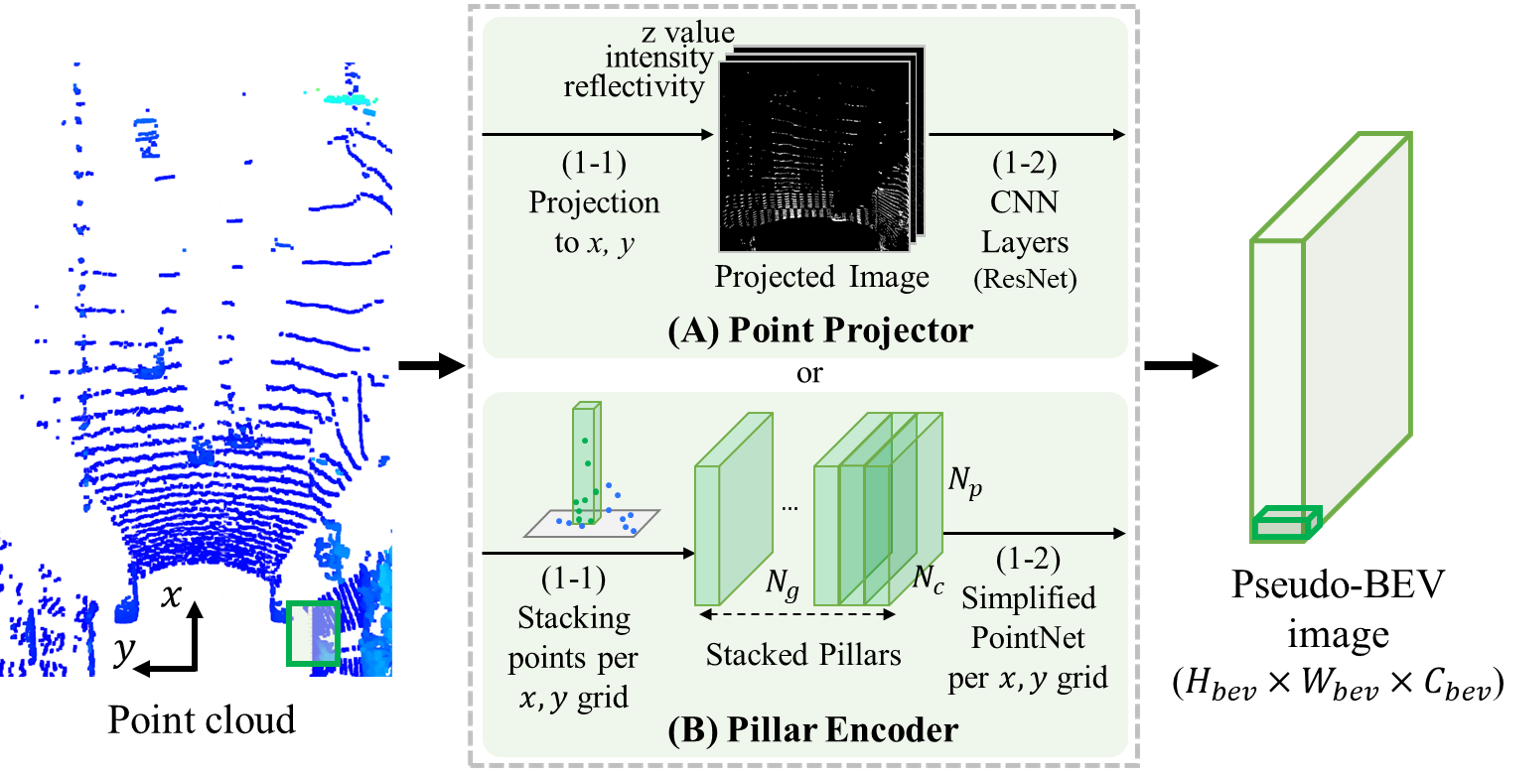}
\caption{Detail structure of two BEV encoder: Point Projector and Pillar Encoder.}
\label{enc_detail}
}
\end{figure}

{\noindent\textbf{Dataset Composition.}} The K-lane is divided into fifteen directories, each representing a sequence. Each directory has one associated file that describe the driving condition of the frames in the sequence, and contains files for the collected point cloud data, BEV point cloud tensor (i.e., stacked pillars shown in Fig. \ref{enc_detail}), BEV label, front (camera) images, and calibration parameters, as shown in Table \ref{tab:a_3}. Pedestrians’ faces are blurred on the front images for privacy protection. Interface for pre-processing the files is provided in Section A.2.

\begin{table}[htb!]
{
\centering
\small
\tabcounterint
\begin{tabular}{cccc}
\hlineB{3}
\begin{tabular}[c]{@{}c@{}}Seq-\\ uence\end{tabular} & \begin{tabular}[c]{@{}c@{}}Num. \\ Frames\end{tabular} & Location & Time  \\ \hlineB{2}
1 & 1708 & Urban roads {[}Sejong{]} & Night \\
2 & 803 & Urban roads {[}Daejeon{]} & Day   \\
3 & 549  & Urban roads {[}KAIST{]} & Day   \\
4 & 1468 & Urban roads {[}KAIST{]} & Day   \\
5 & 251 & Urban roads {[}Daejeon{]} & Day   \\
6 & 132 & Urban roads {[}Daejeon{]} & Day   \\
7 & 388 & Urban roads {[}Daejeon{]} & Day   \\
8 & 357 & Urban roads {[}Daejeon{]} & Day   \\ 
9 & 654 & Urban roads {[}Daejeon{]}  & Day   \\
10 & 648 & Urban roads {[}Daejeon{]}  & Day   \\
11 & 1337 & Urban roads {[}Daejeon{]}  & Night \\
12 & 370 & Urban roads {[}Daejeon{]}  & Night \\
13 & 2991 & Highway {[}Daejeon to Cheongju{]} & Day   \\
14 & 1779 & Highway {[}Daejeon to Cheongju{]} & Night \\
15 & 1947 & Highway {[}Cheongju to Daejeon{]} & Night \\ \hlineB{3}
\end{tabular}
\caption{Sequences in K-Lane.}
\label{tab:sequence}
}
\end{table}

\begin{table*}[htb!]
{
\tabcounterint
\small
\centering
\begin{tabular}{c|l|c}
\hlineB{3}
Conditions & \multicolumn{1}{c|}{Explanation} & Num. Frames \\ \hlineB{3}
Urban & Data acquired from city or university & 8607 \\ \hline
Highway & Data acquired on Highway & 6775 \\ \hline
Night & Data acquired at night (approximately 20:00-2:00) & 7139 \\ \hline
Daytime & Data acquired during the daytime (about 12:00-16:00) & 8243 \\ \hline
Normal & Data without curved or merging lanes (mostly straight lanes) & 11065 \\ \hline
Gentle Curve & Data with curved lanes whose radius of curvature is greater than 160 {[}m{]} & 1804 \\ \hline
Sharp Curve & Data with curved lanes whose radius of curvature is less than 160 {[}m{]} & 1431 \\ \hline
Merging & Data with a converging or diverging lane at the rightmost or leftmost lane & 982 \\ \hline
No Occlusion & Data without occluded lanes based on the lane label & 9443 \\ \hline
Occlusion 1 & Data with one occluded lane based on the lane label & 2660 \\ \hline
Occlusion 2 & Data with two occluded lanes based on the lane label & 2112 \\ \hline
Occlusion 3 & Data with three occluded lanes based on the lane label & 793 \\ \hline
Occlusion 4-6 & \begin{tabular}[c]{@{}l@{}}Data with four to six occluded lanes based on the lane label; \\ Since there are few samples of data with five or six occluded lanes, \\ they are integrated as a single condition (i.e., four to six occluded lanes).\end{tabular} & 374 \\ \hlineB{3}
\end{tabular}
\caption{Condition details}
\label{tab:a_4}
}
\end{table*}

\begin{table*}[htb!]
{
\tabcounterint
\small
\centering
\begin{tabular}{c|c|l|l}
\hlineB{3}
Datum Type & Extension & \multicolumn{1}{c|}{Format} & \multicolumn{1}{c}{Comment} \\ \hlineB{2}
Point cloud & .pcd & Point cloud with 131072 points & \begin{tabular}[c]{@{}l@{}}Input to point projector and \\ heuristic technique\end{tabular} \\ \hline
BEV point cloud tensor & .pickle & $N_g \times N_c \times N_p$ size array & Input to pillar encoder \\ \hline
BEV label & .pickle & $H_{bev} \times (W_{bev}+6)$ size array & \begin{tabular}[c]{@{}l@{}}Lane label including unlabeled lane \\ per row (6 columns are for the possible \\ row-wise detection-based approaches)\end{tabular} \\ \hline
Front image & .png & RGB image & For annotation and visualization \\ \hline
Calibration parameters & .txt & Intrinsic and extrinsic parameters & For Lidar-camera projection \\ \hline
Condition & .txt & Condition (e.g., night and day) & For evaluation \\ \hlineB{3}
\end{tabular}
\caption{Dataset Composition}
\label{tab:a_3}
}
\end{table*}

\vspace{2mm}
{\large\noindent\textbf{A.2. Details of Development Kits}}
\vspace{2mm}

In addition to the K-Lane dataset, we also provide the devkits which can be used to expand the dataset, and to develop further LLDNs.
The devkits are available to the public in the form of three programs: (1) TPC - Total Pipeline Code for training and evaluation, (2) GLT - Graphic User Interface (GUI)-based Labeling Tools, and (3) GDT – GUI-based development Tools for evaluation, visualization, and additional conditions annotations. 

{\noindent\textbf{Total Pipeline Code.}} TPC is a complete neural network development code that supports pre-processing of the input data and label, train the network, and perform evaluation based on the F1-metric.
TPC handles input and output as Python dictionaries and support modularization of the LLDN (BEV encoder, GFC, detection head), therefore, providing comprehensive and flexible support to developers.

\begin{figure*}[htb!]
{
\figcounterint
\centering
\includegraphics[width=1.9\columnwidth]{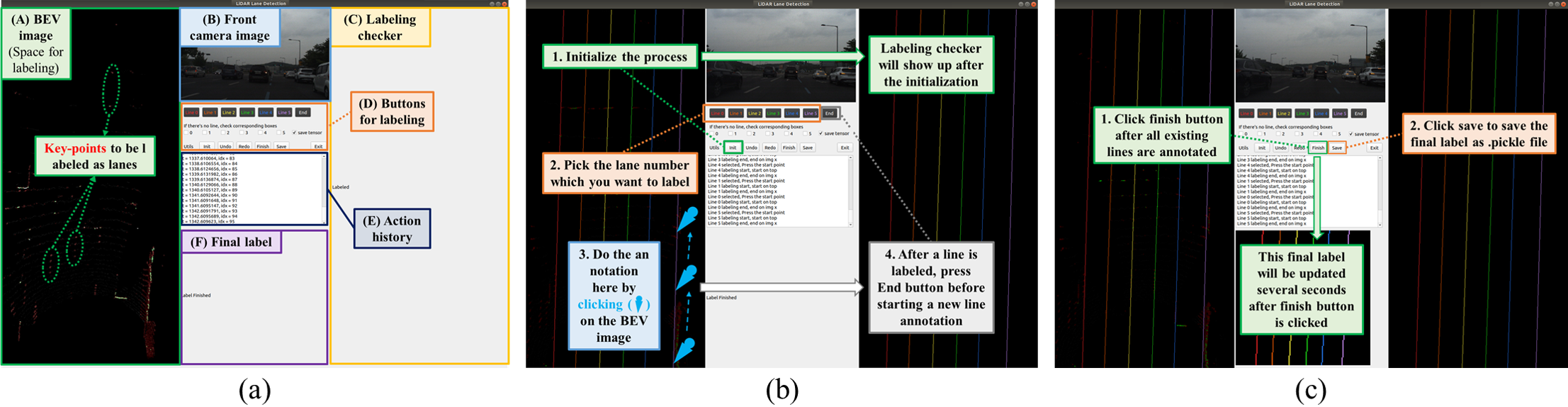}
\caption{GUI-based Labeling Tool (GLT): (a) Overall components of GLT, (b) Labeling process of a point cloud, (c) Finalizing and saving the label.}
\label{glt}
}
\end{figure*}

{\noindent\textbf{GUI-based Labeling Tools.}} GLT provides an easy way to develop a labeled dataset for a Lidar and a front view camera, regardless of the Lidar and camera models. As shown in Fig. \ref{glt} (left), GLT provides an easy way for labeling by showing the intensity of point cloud in a BEV image. Fig. \ref{glt} (middle) shows a synchronized front camera image for easy labeling of point cloud, and Fig. \ref{glt} (right) shows the saved labeled point cloud.

\begin{figure*}[htb!]
{
\figcounterint
\centering
\includegraphics[width=2\columnwidth]{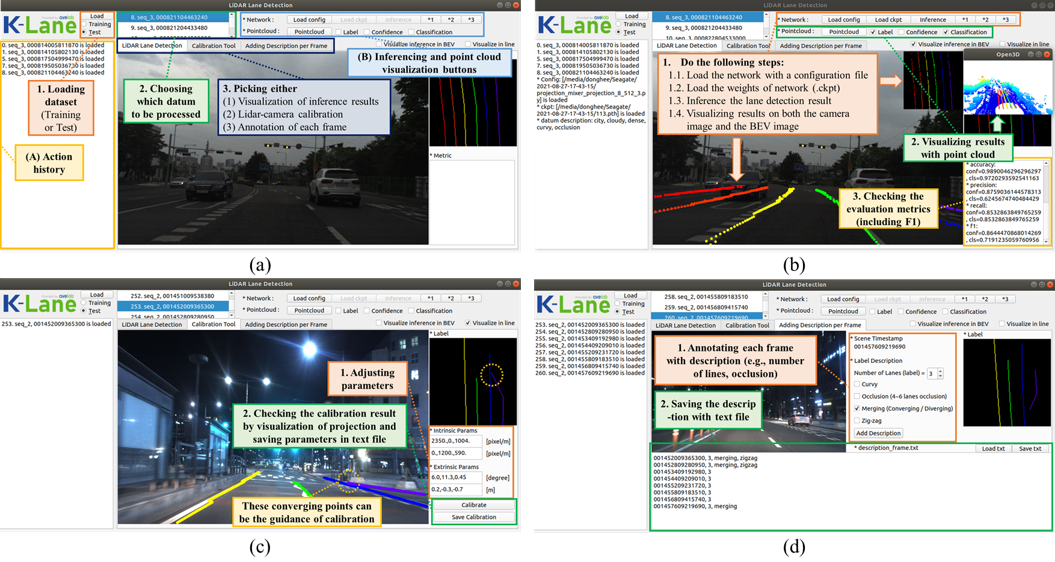}
\caption{GUI-based Development Tools (GDT): (a) overall components of GDT and loading a data, (b) visualization of the LLDN inference results, (c) calibrating Lidar with camera (d) annotating a frame.}
\label{gdt}
}
\end{figure*}

{\noindent\textbf{GUI-based Development Tools.}} GDT is a GUI program used together with TPC. GDT provides visualization of inference results for each scene as point cloud or camera image with projected lanes (Fig. \ref{gdt}-b), high-accuracy calibration of camera and Lidar sensors with specific points of the lanes (Fig. \ref{gdt}-c), and annotation of each frame with set of buttons (Fig. \ref{gdt}-d).

\vspace{5mm}
{\large\noindent\textbf{B. Details of LLDNs}}
\vspace{2mm}

This section provides a detailed neural network structure of the LLDN-GFC proposed in Section 3.2 of the main paper and its counterparts, CNN-based LLDN.

\vspace{2mm}
{\large\noindent\textbf{B.1. Details of LLDN-GFC}}
\vspace{2mm}

\begin{figure}[b!]
    \figcounterint
    \centering
    \includegraphics[width=0.9\columnwidth]{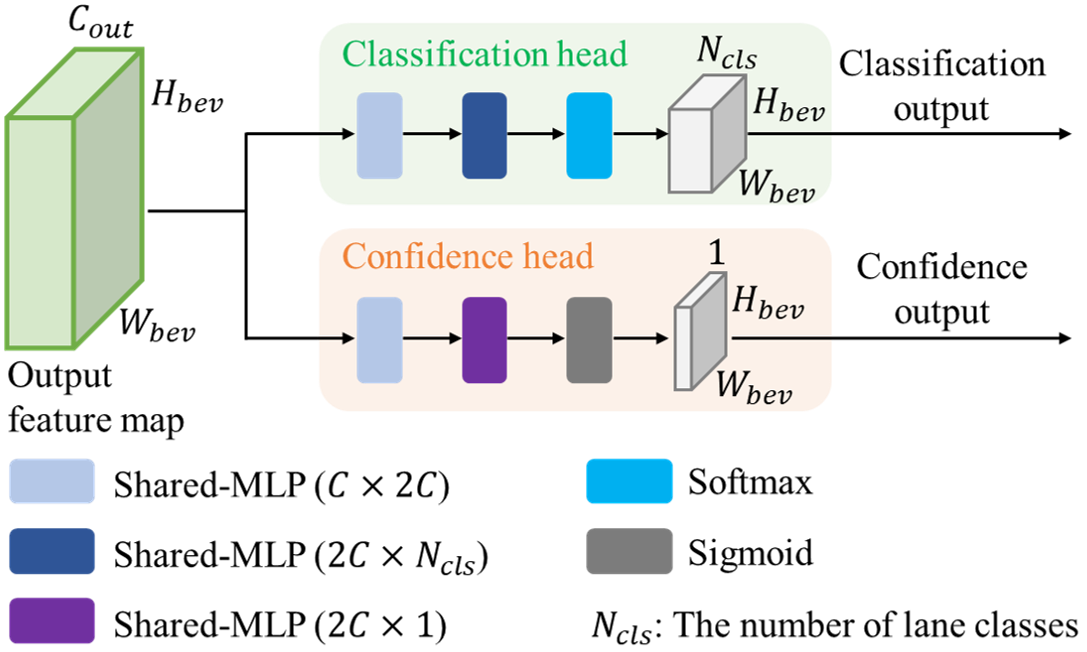}
    \caption{Detection head of the proposed LLDN-GFC.}
    \label{det_structure}
\end{figure}

This section describes the sub-structure of the proposed baseline LLDN-GFC, first shown in Fig. \ref{lldn_structure}.
We divide the LLDN-GFC structure into three parts: the BEV encoder, the global feature corrector (GFC), and the detection head.
The functions (1)$\sim$(5) of Fig. \ref{lldn_structure} are equivalent to the functions (1)$\sim$(5) of Fig. \ref{enc_detail}$\sim$\ref{det_structure}. (e.g., (1) of Fig. \ref{lldn_structure} is equivalent to (1-1) and (1-2) of Fig. \ref{enc_detail}.)

{\noindent\textbf{BEV Encoder.}} BEV encoder projects 3D point cloud into a horizontal plane to produce 2D pseudo-BEV image. A large number of heuristic path planning algorithms, such as A* \cite{astar}, RRT* \cite{rrt}, and End-to-End autonomous driving algorithms \cite{learningbycheating} require lane lines on 2D BEV images. The proposed LLDN-GFC variants use one of the two most common 2D BEV encoders, as shown in Fig. \ref{enc_detail}.

The primary 2D BEV encoder for the LLDN-GFC is the point projector \cite{complex_yolo,joint_projector} that projects point clouds into xy-horizontal plane and produces pseudo-BEV images using CNN. In this case, three additional information (z, intensity, and reflectivity) other than x and y of the point cloud is used to generate three channels of the produced pseudo-BEV image. In order not to lose lane information while maintaining the real-time speed, we use only to a depth of the CNN where the feature map becomes the $1$/$8^2$ of the pseudo-BEV image input. To this end, we may use the first 14, 28, and 41 convolutional layers of the ResNet-18, ResNet-34, and ResNet-50 \cite{resnet}, respectively. Note that we denote these partial ResNets as ResNet14, ResNet28, and ResNet41 in the ablation studies in Section C, and that ResNet28 is the one used for the proposed LLDN-GFC.

An alternative for low computational 2D BEV encoder is the pillar encoder based on Point Pillars \cite{pillars} that has relatively small network size. Pillar encoder has slightly lower performance but faster inference speed than the CNN-based point projector. Therefore, in this paper, pillar encoder is presented for real-time applications. As shown in Fig. \ref{enc_detail}, the pillar encoder aligns the point cloud in each grid of the horizontal plane to generate stacked pillars of size ${N_g}\times{N_c}\times{N_p}$, where ${N_g}$ is the total number of grids, ${N_c}$ is the point feature components, and ${N_p}$ is the maximum number of points present on the grid. Then, a simplified version of PointNet \cite{pointnet} consisting of shared MLP’s of size ${N_c}\times{C}$ is applied to each grid to extract pseudo-BEV image of size ${{H_{bev}}\times{W_{bev}}\times{C}}$. In this paper, considering that a lane in the real-world has a width of about 16cm and stretches long in the longitudinal direction of the road, the grid size in the pseudo-BEV is set to 32cm in the longitudinal direction and 16cm in the lateral direction.

{\noindent\textbf{Global Feature Correlator.}} Due to the advantage of patch-wise self attention networks (i.e., calculating the correlation in high resolution between distant grids within the feature map) for Lidar lane detection, we utilize two types of patch-wise self-attention network for global correlation, ViT \cite{vit} and MLP-Mixer \cite{mixer} to propose two possible GFCs, GFC-T (the main proposal) and GFC-M (the computational alternative), respectively. In this section, we provide the structure of those GFCs in detail.

\begin{figure}[htb!]
{
\figcounterint
\centering
\includegraphics[width=1.0\columnwidth]{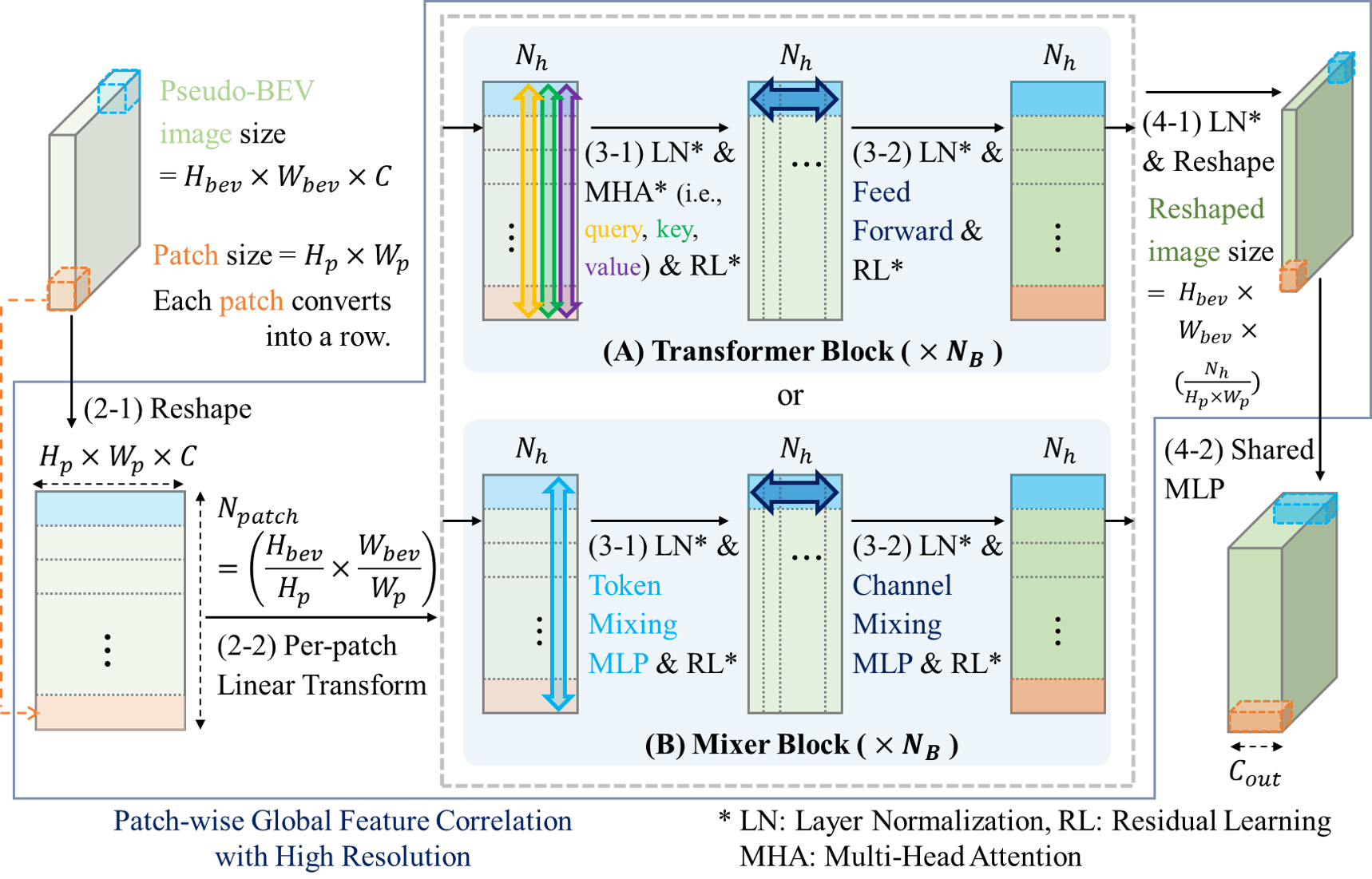}
\caption{Details of proposed Global Feature Correlators; the input size is expressed with height $H_{bev}$, width $W_{bev}$, and number of channels $C$, and a patch size has height $H_p$ and width $W_p$. The input and output size in Mixer block and Transformer encoder block is the same and the block repeats $N_D$ times.}
\label{gfc_detail}
}
\end{figure}

Fig. \ref{gfc_detail} shows the details of the two types of GFC, GFC-T and GFC-M. Both of the two GFCs employ (2-1), (2-2), (4-1), and (4-2) functions, while GFC-T employs (A) Transformer blocks and, a low computational alternative, GFC-M uses (B) Mixer blocks for global correlation. In Fig. \ref{gfc_detail}, function (2-1) reshapes the pseudo-BEV image into a 2D tensor for global correlation. Function (2-2) performs per-patch linear transform, and functions (3-1) and (3-2) perform global correlation through per-channel MLPs (i.e., Multi-head attention or Token-mixing MLP) and per-patch MLPs (i.e., Feed forward or Channel-mixing MLP), respectively. The Transformer encoder block in (A) performs global correlation between image patches using three MLPs calculating query, key, and value and utilizes the global correlation result to pay more attention (i.e., larger attention score) to the important patches to improve the global feature extraction. In addition, the Transformer encoder block allows visualization of the attention score, which can be used for analyzing the network inference, as shown in Section 4.2. However, since the Transformer encoder block becomes a large network for the three MLPs and repeats calculating the attention score for every query (i.e., patch), the total computational cost increases in quadratic with the number of patches. On the other hand, Mixer block in (B) replace the multi-head attention, (3-1) in (A), with a single MLP, (3-1) in (B), which allows smaller network size and lower computational cost but it becomes difficult to analyze the network through attention score and causes lower model inductive bias than the Transformer block. Nonetheless, the two types of GFCs (GFC-T and GFC-M) show strong performance improvement in the Lidar lane detection.

Function (4-1) reshapes the last output of Transformer encoder and Mixer block to the size required for the lane detection head. Note that ViT and MLP-Mixer for the conventional image classification compress the detected feature with a classification token and global average pooling, respectively, but the proposed GFC reshapes the size up to ${H_{bev}}\times{W_{bev}}$ that is the input size to the function (2-1). This is how the proposed GFC provides inductive bias to the output feature map, which is testified with the visualized heatmap in Section 4.1, where high activation result is obtained in the resolution of pixels (much smaller resolution than patches). Note that the number of total pixels after the reshape becomes ${H_p}\times{W_p}$ times the number of total patches (${N_{patch}}={H_{bev}}/{{H_p}\times{W_{bev}}/{W_p}}$) before the reshape, which means that each pixel of the reshaped feature map has ${N_h}/({H_p}\times{W_p})$ dimension as a result. Since the channel size of the reshaped output image depends on the hidden dimension $N_h$, it can be smaller than that of the input BEV image, $C_{bev}$. This may cause bottleneck \cite{bottleneck}, so function (4-2) applies 1x1 convolution and produces the final output feature map for the detection head.

{\noindent\textbf{Detection Head.}} Fig. \ref{det_structure} shows the detection head introduced in Section 3.2. There are two segmentation heads: the classification head and the confidence head. Given an input of $H_{bev} \times W_{bev} \times C_{out}$ feature map from the GFC, we employ two sequential shared-MLPs to create the final prediction maps output. The first shared-MLP expands the dimension of the feature map from $C_{out}$ to $2C_{out}$ for both classification and confidence heads to increase the representation capacity. The second shared-MLP then transforms the feature maps from $2C_{out}$ to $N_{cls}$ and from $2C_{out}$ to $1$ for the classification head and confidence head, respectively, resulting in a classification map and confidence map predictions. We then apply a grid-wise softmax to the classification map to get the $H_{bev} \times W_{bev} \times N_{cls}$ classification map output, and a grid-wise sigmoid to the confidence map to get the $H_{bev} \times W_{bev} \times 1$ confidence map output. The classification map shows per-class-probabilities of each grid, while the confidence map only shows the probability of the grid being a lane or not. The implementation of both classification and confidence tasks in parallel enables the LLDN to simultaneously predict the lane shape and the lane class.

As stated in Section 3.2, we use the soft dice loss \cite{diceloss} for supervising the confidence loss $\mathcal{L}_{conf}$, defined as
{
\small
\begin{equation}
    \mathcal{L}_{conf} = 1 - \frac{2 \sum_i^N \sum_j^M x_{conf_{i,j}} \hat{x}_{conf_{i,j}}}{\sum_i^N \sum_j^M x_{conf_{i,j}}^2 + \sum_i^N \sum_j^M \hat{x}_{conf_{i,j}}^2 + \epsilon},
    \label{softdice_eq}
\end{equation}
\normalsize
}
\noindent where $\epsilon$ is set to be $10^{-12}$ to prevent division by zero. The grid-wise cross-entropy loss \cite{unet} is used as the classification loss $\mathcal{L}_{cls}$, defined as

\begin{equation}
    \mathcal{L}_{cls} = \frac{1}{NM}\sum_i^N\sum_j^M log(p(\hat{\bm{x}}_{cls_{i,j}})),
    \label{cross-entropy_eq}
\end{equation}

\noindent where $p(\hat{\bm{x}}_{cls_{i,j}})$ is the softmax of the classification prediction for class $k = x_{cls_{i,j}}$ at grid $(i,j)$, defined as

\begin{equation}
    p(\hat{\bm{x}}_{cls_{i,j}}) = \frac{exp(\hat{x}_{cls_{i,j,k}})}{\sum_{k'=1}^C exp(\hat{x}_{cls_{i,j,k'}})}.
    \label{softmax_eq}
\end{equation}

The grid-wise cross-entropy loss penalizes the network based on the deviation of $ p(\hat{\bm{x}}_{cls_{i,j}})$ from 1, which is equivalent to maximizing the probability of the correct class for each grid on the final classification map output. The total loss function $\mathcal{L}_{total}$ is the summation of both classification loss and the confidence loss as
\begin{equation}
    \mathcal{L}_{total} = \mathcal{L}_{conf} + \mathcal{L}_{cls}.
    \label{total_loss_eq}
\end{equation}

\vspace{2mm}
{\large\noindent\textbf{B.2. Details of CNN-based LLDN}}
\vspace{2mm}

\begin{figure}[b!]
\centering
\includegraphics[width=0.8\columnwidth]{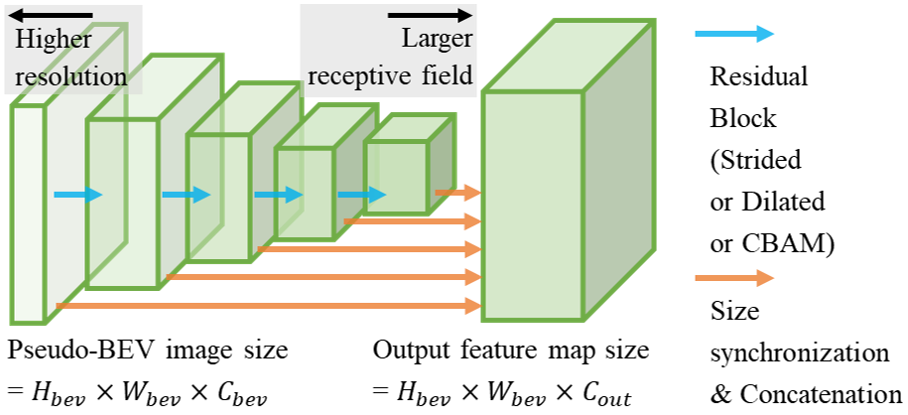}
\caption{Overall structure of the conventional CNN-based backbone.}
\label{cnn_fpn}
\end{figure}

As we introduce in Section 4.1, we consider three types of CNN-based backbone as the counterparts of GFC, ResNet and FPN (RNF)-based backbone: (1) RNF-S, (2) RNF-D, and (3) RNF-C, where S, D, and C represent residual blocks implemented with strided convolution, dilated convolution, and convolutional block attention module (CBAM) \cite{cbam}, respectively. As shown in Fig. \ref{cnn_fpn}, there are 5 residual blocks composed of 3, 5, 5, 5, and 5 convolutional layers in the ResNet side. Each block produces a feature map that is $2^2$ times smaller than the input feature map, and the feature pyramid network (FPN) concatenates the feature maps from each block to produce the final output feature map.

\vspace{5mm}
{\large\noindent\textbf{C. Ablations}}
\vspace{2mm}

In this section, we perform several ablation studies on the proposed LLDN-GFC, the low computational alternative, and the the conventional CNN-based LLDNs.

\begin{table*}[htb!]
{
\tabcounterint
\small
\centering
\begin{tabular}{c|c|c|c|c|c|c|c|c|c|c|c|c|c|c|c}
\hlineB{3}
{\begin{tabular}[c]{@{}c@{}}Back-\\ bone\end{tabular}} & {Total} & {Day} & {Night} & {\begin{tabular}[c]{@{}c@{}}Ur-\\ ban\end{tabular}} & {\begin{tabular}[c]{@{}c@{}}High-\\      way\end{tabular}} & {\begin{tabular}[c]{@{}c@{}}Nor-\\ mal\end{tabular}} & {\begin{tabular}[c]{@{}c@{}}Gentle \\ Curve\end{tabular}} & {\begin{tabular}[c]{@{}c@{}}Sharp\\ Curve\end{tabular}} & {\begin{tabular}[c]{@{}c@{}}Mer-\\ ging\end{tabular}} & {\begin{tabular}[c]{@{}c@{}}No\\ Occ\end{tabular}} & {\begin{tabular}[c]{@{}c@{}}Occ\\ 1\end{tabular}} & {\begin{tabular}[c]{@{}c@{}}Occ\\ 2\end{tabular}} & {\begin{tabular}[c]{@{}c@{}}Occ\\ 3\end{tabular}} & {\begin{tabular}[c]{@{}c@{}}Occ\\ 4-6\end{tabular}} & {FPS} \\ \hlineB{2}
\multirow{2}{*}{\begin{tabular}[c]{@{}c@{}}GFC\\ -T1\end{tabular}} & 77.5 & 77.3 & 77.8 & 76.1 & 79.1 & 78.2 & 78.1 & 70.3 & 77.3 & 78.5 & 76.9 & 76.5 & 73.1 & 69.8 &  \multirow{2}{*}{12.7} \\
 & 76.1 & 76.1 & 76.1 & 74.6 & 77.9 & 76.9 & 76.6 & 68.4 & 76.5 & 77.1 & 74.9 & 75.7 & 73.1 & 69.0 &  \\ \hline
\multirow{2}{*}{\begin{tabular}[c]{@{}c@{}}GFC\\ -T3\end{tabular}} & \textbf{81.0} & \textbf{81.0} & \textbf{80.9} & \textbf{80.1} & \textbf{82.0} & \textbf{82.0} & \textbf{81.7} & \textbf{75.2} & \textbf{79.8} & \textbf{81.8}& \textbf{80.5} & \textbf{80.1} & \textbf{77.2} & \textbf{75.7} & \multirow{2}{*}{12.6} \\
 & \textbf{80.1} & \textbf{80.4} & \textbf{79.7} & \textbf{79.0} & \textbf{81.3} & \textbf{81.0} & \textbf{81.0} & \textbf{74.0} & \textbf{79.1} & \textbf{80.8} & \textbf{79.2} & \textbf{79.8} & \textbf{76.4} & \textbf{75.4} &  \\ \hline
\multirow{2}{*}{\begin{tabular}[c]{@{}c@{}}GFC\\ -M1\end{tabular}} & 74.8 & 74.7 & 74.9 & 72.9 & 77.1 & 74.2 & 77.0 & 66.1 & 73.1 & 75.5 & 74.5 & 75.2 & 70.6 & 65.0 & \multirow{2}{*}{16.1} \\
 & 73.4 & 73.6 & 73.3 & 71.2 & 76.1 & 74.2 & 75.7 & 63.7 & 72.2 & 74.1 & 72.2 & 74.8 & 69.3 & 64.1 &  \\ \hline
\multirow{2}{*}{\begin{tabular}[c]{@{}c@{}}GFC\\ -M3\end{tabular}} & 78.9 & 79.0 & 78.9 & 77.5 & 80.6 & 79.5 & 80.4 & 72.2 & 77.6 & 79.8 & 78.3 & 78.5 & 74.7 & 70.1 & \multirow{2}{*}{15.4}  \\
 & 77.8 & 78.0 & 77.6 & 76.2 & 79.8 & 78.4 & 79.4 & 70.5 & 76.7 & 78.7 & 76.8 & 77.8 & 73.8 & 69.4 &  \\ \hline
\multirow{2}{*}{\begin{tabular}[c]{@{}c@{}}{RNF}\\ -S8\end{tabular}} & 74.6 & 73.3 & 76.0 & 73.2 & 76.2 & 74.8 & 76.1 & 69.6 & 75.8 & 76.5 & 73.9 & 71.7 & 66.3 & 61.8 &  \multirow{2}{*}{\textbf{16.5}} \\
 & 58.0 & 58.0 & 58.0 & 58.8 & 57.1 & 58.3 & 57.3 & 54.9 & 62.0 & 60.8 & 55.0 & 54.2 & 50.3 & 42.6 &  \\ \hline
\multirow{2}{*}{\begin{tabular}[c]{@{}c@{}}{RNF}\\ -C8\end{tabular}} & 77.7 & 76.4 & 79.3 & 76.3 & 79.4 & 78.0 & 79.6 & 72.2 & 78.0 & 79.5 & 77.0 & 75.0 & 70.7 & 66.9 &  \multirow{2}{*}{15.5}\\
 & 63.7 & 62.5 & 65.0 & 62.6 & 65.0 & 63.9 & 64.7 & 58.1 & 65.9 & 66.3 & 60.9 & 59.6 & 56.7 & 48.6 &  \\ \hline
\multirow{2}{*}{\begin{tabular}[c]{@{}c@{}}{RNF}\\ -D8\end{tabular}} & 74.8 & 73.4 & 76.5 & 73.1 & 77.0 & 74.8 & 77.6 & 70.1 & 75.9 & 76.7 & 74.6 & 72.2 & 65.9 & 62.0 & \multirow{2}{*}{15.3} \\
 & 55.4 & 55.5 & 55.4 & 56.1 & 54.6 & 55.5 & 55.3 & 52.7 & 60.7 & 57.7 & 53.1 & 53.0 & 47.4 & 42.1 &  \\ \hline
\multirow{2}{*}{\begin{tabular}[c]{@{}c@{}}RNF\\ -S13\end{tabular}} & 67.4 & 65.9 & 69.2 & 66.4 & 68.7 & 67.3 & 69.6 & 63.9 & 69.6 & 69.2 & 67.0 & 64.3 & 59.6 & 56.8 & \multirow{2}{*}{15.0} \\
 & 62.0 & 60.9 & 63.3 & 61.4 & 62.7 & 61.8 & 64.1 & 59.1 & 65.6 & 63.9 & 61.0 & 59.0 & 55.2 & 50.2 &  \\ \hline
\multirow{2}{*}{\begin{tabular}[c]{@{}c@{}}RNF\\ -C13\end{tabular}} & 78.0 & 77.1 & 79.0 & 77.1 & 79.2 & 78.4 & 79.2 & 72.2 & 78.7 & 79.5 & 77.4 & 76.1 & 71.6 & 66.6 & \multirow{2}{*}{14.9} \\
 & 69.2 & 69.2 & 69.3 & 68.5 & 70.2 & 69.7 & 70.4 & 62.7 & 70.6 & 71.2 & 67.1 & 67.4 & 63.0 & 54.6 &  \\ \hline
\multirow{2}{*}{\begin{tabular}[c]{@{}c@{}}RNF\\ -D13\end{tabular}} & 76.9 & 75.7 & 78.2 & 75.5 & 78.5 & 77.0 & 79.0 & 71.5 & 77.8 & 78.7 & 76.2 & 74.3 & 69.1 & 63.6 & \multirow{2}{*}{14.8} \\
 & 60.4 & 60.2 & 60.7 & 61.7 & 58.9 & 60.8 & 59.7 & 56.21 & 65.5 & 62.9 & 57.6 & 57.1 & 53.8 & 46.7 &  \\ \hlineB{3}
\end{tabular}
\caption{\label{tab:e_1}Proj14-based LLDN performance for backbones with various depth.}
}
\end{table*}

\begin{table*}[htb!]
{
\tabcounterint
\small 
\centering
\begin{tabular}{c|c|c|c|c|c|c|c|c|c|c|c|c|c|c|c}
\hlineB{3}
{\begin{tabular}[c]{@{}c@{}}Back-\\ bone\end{tabular}} & {Total} & {Day} & {Night} & {\begin{tabular}[c]{@{}c@{}}Ur-\\ ban\end{tabular}} & {\begin{tabular}[c]{@{}c@{}}High-\\ way\end{tabular}} & {\begin{tabular}[c]{@{}c@{}}Nor-\\ mal\end{tabular}} & {\begin{tabular}[c]{@{}c@{}}Gentle \\ Curve\end{tabular}} & {\begin{tabular}[c]{@{}c@{}}Sharp \\ Curve\end{tabular}} & {\begin{tabular}[c]{@{}c@{}}Mer-\\ ging\end{tabular}} & {\begin{tabular}[c]{@{}c@{}}No \\ Occ\end{tabular}} & {\begin{tabular}[c]{@{}c@{}}Occ\\ 1\end{tabular}} & {\begin{tabular}[c]{@{}c@{}}Occ\\ 2\end{tabular}} & {\begin{tabular}[c]{@{}c@{}}Occ\\ 3\end{tabular}} & {\begin{tabular}[c]{@{}c@{}}Occ\\ 4-6\end{tabular}} & {FPS} \\ \hlineB{2}
\multirow{2}{*}{\begin{tabular}[c]{@{}c@{}}GFC\\ -T1\end{tabular}} & 79.8 & 79.6 & 80.0 & 79.4 & 80.3 & 80.2 & 80.6 & 75.2 & 79.4 & 80.8 & 79.1 & 78.7 & 74.9 & 72.8 & \multirow{2}{*}{11.8} \\
 & 78.8 & 78.9 & 78.7 & 78.3 & 79.4 & 79.3 & 75.2 & 73.5 & 78.4 & 79.8 & 77.7 & 78.2 & 73.9 & 72.5 &  \\ \hline
\multirow{2}{*}{\begin{tabular}[c]{@{}c@{}}GFC\\ -T3\end{tabular}} & \textbf{82.1} & \textbf{82.2} & \textbf{82.0} & \textbf{81.7} & \textbf{82.5} & \textbf{82.5} & 82.2 & \textbf{78.0} & \textbf{81.0} & \textbf{82.9} & \textbf{81.4} & \textbf{82.3} & \textbf{78.7} & 75.9 & \multirow{2}{*}{11.6} \\
 & \textbf{81.1} & \textbf{81.4} & \textbf{80.7} & \textbf{80.6} & \textbf{81.7} & \textbf{81.5} & \textbf{83.0} & \textbf{76.7} & \textbf{80.1} & \textbf{81.9} & \textbf{81.4} & \textbf{81.3} & \textbf{78.7} & 75.5 &  \\ \hline
\multirow{2}{*}{\begin{tabular}[c]{@{}c@{}}GFC\\ -T5\end{tabular}} & 81.1 & 81.0 & 81.2 & 80.6 & 81.7 & 82.0 & \textbf{82.3} & 76.0 & 80.0 & 82.1 & 80.5 & 79.6 & 77.3 & \textbf{77.2} & \multirow{2}{*}{11.2} \\
 & 79.5 & 79.5 & 79.4 & 78.7 & 80.4 & 80.0 & 80.7 & 73.0 & 78.8 & 80.3 & 78.5 & 78.6 & 75.3 & \textbf{76.2} &  \\ \hline
\multirow{2}{*}{\begin{tabular}[c]{@{}c@{}}GFC\\ -M1\end{tabular}} & 78.5 & 78.5 & 78.4 & 77.8 & 79.3 & 78.9 & 80.0 & 72.5 & 78.0 & 79.4 & 77.8 & 77.7 & 74.5 & 70.2 & \multirow{2}{*}{\textbf{13.4}} \\
 & 77.3 & 77.6 & 77.0 & 76.4 & 78.4 & 77.8 & 79.1 & 70.2 & 76.9 & 78.2 & 77.8 & 77.7 & 74.5 & 69.5 &  \\ \hline
\multirow{2}{*}{\begin{tabular}[c]{@{}c@{}}GFC\\ -M3\end{tabular}} & 79.7 & 79.9 & 79.6 & 78.9 & 80.8 & 80.1 & 81.3 & 74.6 & 79.0 & 80.4 & 79.6 & 79.4 & 76.1 & 72.5 & \multirow{2}{*}{13.3} \\
 & 78.8 & 79.0 & 78.4 & 77.7 & 80.0 & 79.2 & 80.4 & 72.6 & 78.1 & 79.4 & 78.3 & 78.9 & 74.9 & 71.7 &  \\ \hline
\multirow{2}{*}{\begin{tabular}[c]{@{}c@{}}GFC\\ -M5\end{tabular}} & 78.7 & 77.3 & 78.8 & 78.0 & 79.6 & 79.0 & 80.5 & 73.5 & 77.6 & 79.6 & 78.2 & 78.1 & 74.7 & 69.9 & \multirow{2}{*}{13.1} \\
 & 79.2 & 79.5 & 78.9 & 78.4 & 80.1 & 79.0 & 81.1 & 73.6 & 78.4 & 80.1 & 78.6 & 78.6 & 75.3 & 71.4 &  \\ \hline
\multirow{2}{*}{\begin{tabular}[c]{@{}c@{}}{RNF}\\ -S8\end{tabular}} & 74.6 & 73.9 & 75.4 & 74.4 & 74.9 & 74.9 & 75.3 & 70.5 & 76.0 & 76.5 & 73.5 & 71.8 & 66.9 & 64.8 & \multirow{2}{*}{13.2} \\
 & 63.0 & 62.8 & 63.3 & 63.4 & 62.6 & 63.5 & 62.7 & 57.6 & 66.7 & 65.3 & 60.3 & 60.9 & 54.7 & 49.8 &  \\ \hline
\multirow{2}{*}{\begin{tabular}[c]{@{}c@{}}{RNF}\\ -C8\end{tabular}} & 78.1 & 77.3 & 79.1 & 77.6 & 78.7 & 78.2 & 79.7 & 74.9 & 79.0 & 79.7 & 77.3 & 76.1 & 71.5 & 68.6 & \multirow{2}{*}{13.1} \\
 & 70.3 & 69.7 & 71.0 & 69.8 & 70.9 & 70.4 & 71.6 & 66.6 & 72.0 & 72.2 & 68.0 & 69.0 & 62.8 & 58.5 &  \\ \hline
\multirow{2}{*}{\begin{tabular}[c]{@{}c@{}}RNF\\ -S13\end{tabular}} & 73.2 & 72.6 & 74.0 & 73.1 & 73.3 & 73.3 & 74.0 & 70.5 & 74.8 & 74.9 & 72.2 & 70.9 & 65.7 & 63.5 & \multirow{2}{*}{13.1} \\
 & 70.5 & 70.1 & 71.0 & 70.4 & 70.6 & 70.4 & 71.9 & 68.1 & 72.5 & 72.3 & 68.4 & 69.0 & 63.3 & 59.0 &  \\ \hline
\multirow{2}{*}{\begin{tabular}[c]{@{}c@{}}RNF\\ -C13\end{tabular}} & 78.0 & 77.6 & 78.5 & 77.7 & 78.3 & 77.9 & 80.0 & 76.0 & 78.9 & 79.6 & 76.9 & 76.0 & 71.9 & 69.3 & \multirow{2}{*}{13.0} \\
 & 75.3 & 75.1 & 75.5 & 74.8 & 76.0 & 75.0 & 77.9 & 73.1 & 76.5 & 77.0 & 73.1 & 74.1 & 69.2 & 65.3 &  \\ \hline
\multirow{2}{*}{\begin{tabular}[c]{@{}c@{}}RNF\\ -D13\end{tabular}} & 69.5 & 68.4 & 70.8 & 69.5 & 69.6 & 69.6 & 70.1 & 67.3 & 72.1 & 71.6 & 68.5 & 66.2 & 61.5 & 58.5 & \multirow{2}{*}{13.1} \\
 & 65.5 & 64.9 & 66.2 & 65.6 & 65.3 & 65.5 & 65.8 & 62.9 & 68.7 & 67.6 & 63.1 & 63.0 & 58.5 & 54.6 &  \\ \hline
\multirow{2}{*}{\begin{tabular}[c]{@{}c@{}}RNF\\ -D23\end{tabular}} & 72.1 & 71.3 & 73.0 & 71.9 & 72.3 & 72.2 & 72.9. & 69.6 & 74.0 & 74.0 & 70.9 & 69.5 & 63.8. & 61.9 & \multirow{2}{*}{12.7} \\
 & 68.8 & 68.3 & 69.4 & 68.7 & 69.0 & 68.8 & 69.8 & 66.5 & 71.8 & 70.7 & 66.8 & 67.2 & 61.1 & 57.6 &  \\ \hlineB{3}
\end{tabular}
\caption{\label{tab:e_2}Proj28-based LLDN performance for backbones with various depth.}
}
\end{table*}

\noindent \textbf{Ablations on Network Depth.} Since hyperparameters related to the depth of LLDN are BEV encoder depth, the depth of backbone, we provide ablation studies in the following tables (Table \ref{tab:e_1}$\sim$\ref{tab:e_4}) to compare the performance of the LLDN with various BEV encoder, such as Proj14, Proj28, Proj41, and Pillars, and the depth of backbone. Tables in this subsection shows F1-score on the confidence (upper value) and that on the classification (lower value) in each table bin. FPS stands for frame per second, representing the overall computational cost of inference.

As shown in the Table \ref{tab:e_2}, when we use Proj28 and increase the depth of GFC-T from GFC-T1 to GFC-T3, the performance increases by +2.3 and +2.3 in F1-scores on the confidence and the classification, respectively. In addition, as shown in the Table \ref{tab:e_1} and Table \ref{tab:e_2}, when we use GFC-T3 and varies BEV depth from Proj14 to Proj28, the performance increases by +1.1 and +2.2 in F1-scores on the confidence and the classification, respectively. However, performance degradations are observed when the depth of GFC-T is increased from GFC-T3 to GFC-T5 for Proj28 and when depth of BEV encoder is increased from Proj28 to Proj41 for GFC-T3. From our ablation studies, we find that the model with an appropriate capacity can be Proj28-GFC-T3 for the proposed LLDN-GFC, Pillars-GFC-M5 for the low-computational alternative, and Proj28-RNF-S13, Proj28-RNF-C13, Proj28-RNF-D23 for the LLDNs using the conventional CNN-based backbones.

Note that for models with larger capacities, some regularization methods or more sophisticated learning techniques may be applied to reduce overfitting. However, those learning techniques are out of the scope of our study, since we focus on the network architecture and dataset.

\begin{table*}[htb!]
\tabcounterint
{
\small
\centering
\begin{tabular}{c|c|c|c|c|c|c|c|c|c|c|c|c|c|c|c}
\hlineB{3}
{\begin{tabular}[c]{@{}c@{}}Back-\\ bone\end{tabular}} & {Total} & {Day} & {Night} & {\begin{tabular}[c]{@{}c@{}}Ur-\\ ban\end{tabular}} & {\begin{tabular}[c]{@{}c@{}}High-\\      way\end{tabular}} & {\begin{tabular}[c]{@{}c@{}}Nor-\\ mal\end{tabular}} & {\begin{tabular}[c]{@{}c@{}}Gentle \\ Curve\end{tabular}} & {\begin{tabular}[c]{@{}c@{}}Sharp\\ Curve\end{tabular}} & {\begin{tabular}[c]{@{}c@{}}Mer-\\ ging\end{tabular}} & {\begin{tabular}[c]{@{}c@{}}No\\ Occ\end{tabular}} & {\begin{tabular}[c]{@{}c@{}}Occ\\ 1\end{tabular}} & {\begin{tabular}[c]{@{}c@{}}Occ\\ 2\end{tabular}} & {\begin{tabular}[c]{@{}c@{}}Occ\\ 3\end{tabular}} & {\begin{tabular}[c]{@{}c@{}}Occ\\ 4-6\end{tabular}} & {FPS} \\ \hlineB{2}
\multirow{2}{*}{\begin{tabular}[c]{@{}c@{}}GFC\\ -T1\end{tabular}} & 77.8 & 77.7 & 77.9 & 77.5 & 78.2 & 78.3 & 78.5 & 72.3 & 78.0 & 78.9 & 76.7 & 76.5 & 74.2 & 70.0 & \multirow{2}{*}{11.0} \\
 & 76.4 & 76.7 & 76.1 & 76.2 & 76.6 & 77.0 & 76.9 & 70.4 & 77.2 & 77.4 & 75.0 & 75.6 & 73.0 & 69.7 &  \\ \hline
\multirow{2}{*}{\begin{tabular}[c]{@{}c@{}}GFC\\ -T3\end{tabular}} & \textbf{80.5} & \textbf{80.6} & \textbf{80.5} & \textbf{80.4} & 80.7 & \textbf{81.2} & 80.9 & \textbf{75.2} & 79.3 & \textbf{81.4} & \textbf{79.8} & \textbf{79.4} & \textbf{77.3} & \textbf{75.0} & \multirow{2}{*}{10.9} \\
 & \textbf{79.1} & \textbf{79.4} & \textbf{78.8} & \textbf{79.1} & 79.1 & \textbf{79.8} & 79.4 & \textbf{73.4} & 78.3 & \textbf{79.9} & 78.2 & 78.5 & \textbf{75.9} & \textbf{74.5} &  \\ \hline
\multirow{2}{*}{\begin{tabular}[c]{@{}c@{}}GFC\\ -M1\end{tabular}} & 78.2 & 78.3 & 78.1 & 77.0 & 79.7 & 78.8 & 79.6 & 71.5 & 77.2 & 79.0 & 77.6 & 78.1 & 73.9 & 71.2 & \multirow{2}{*}{\textbf{11.7}}  \\
 & 77.1 & 77.4 & 76.7 & 75.7 & 78.9 & 77.8 & 78.4 & 69.4 & 76.3 & 77.9 & 77.5 & 77.5 & 72.7 & 70.6 &  \\ \hline
\multirow{2}{*}{\begin{tabular}[c]{@{}c@{}}GFC\\ -M3\end{tabular}} & 79.9 & 80.1 & 79.7 & 79.2 & \textbf{80.9} & 80.5 & 80.6 & 74.9 & 79.0 & 80.9 & 79.0 & 79.5 & 76.1 & 72.2 & \multirow{2}{*}{11.6} \\
 & 78.8 & 79.2 & 78.4 & 77.8 & \textbf{80.1} & 79.4 & 72.5 & 72.5 & 77.9 & 79.7 & 77.4 & 79.0 & 75.0 & 71.2 &  \\ \hline
\multirow{2}{*}{\begin{tabular}[c]{@{}c@{}}GFC\\ -M5\end{tabular}} & 79.7 & 79.8 & 79.7 & 79.0 & 80.7 & 80.1 & \textbf{81.1} & 74.6 & \textbf{79.6} & 80.6 & 79.4 & 79.1 & 75.7 & 70.6 & \multirow{2}{*}{11.4} \\
 & 78.8 & 79.0 & 78.6 & 77.9 & 79.9 & 79.2 & \textbf{80.3} & 72.7 & \textbf{78.6} & 79.6 & \textbf{78.3} & \textbf{78.7} & 74.8 & 69.9 &  \\ \hline
\multirow{2}{*}{\begin{tabular}[c]{@{}c@{}}RNF\\ -S8\end{tabular}} & 75.1 & 73.8 & 76.5 & 74.4 & 75.8 & 75.0 & 76.3 & 72.3 & 77.1 & 77.0 & 74.6 & 71.8 & 66.8 & 63.6 & \multirow{2}{*}{11.1} \\
 & 61.8 & 61.7 & 62.0 & 62.8 & 60.7 & 61.9 & 61.3 & 59.5 & 67.0 & 64.0 & 59.5 & 59.4 & 54.2 & 48.6 &  \\ \hline
\multirow{2}{*}{\begin{tabular}[c]{@{}c@{}}RNF\\ -C8\end{tabular}} & 76.0 & 75.0 & 77.1 & 75.1 & 77.1 & 76.0 & 77.5 & 72.4 & 77.8 & 77.8 & 74.9 & 73.7 & 68.1 & 65.9 & \multirow{2}{*}{10.7} \\
 & 69.4 & 68.7 & 70.1 & 68.1 & 70.9 & 69.3 & 71.2 & 65.7 & 71.7 & 71.4 & 66.8 & 67.5 & 62.5 & 56.9 &  \\ \hline
\multirow{2}{*}{\begin{tabular}[c]{@{}c@{}}RNF\\ -D8\end{tabular}} & 72.2 & 70.6 & 74.0 & 71.4 & 73.1 & 72.0 & 73.8 & 70.3 & 74.5 & 74.3 & 71.3 & 68.7 & 63.3 & 58.9 & \multirow{2}{*}{10.9} \\
 & 65.9 & 64.7 & 67.2 & 65.3 & 66.6 & 65.5 & 67.7 & 63.8 & 69.3 & 68.1 & 63.8 & 63.1 & 57.6 & 51.5 &  \\ \hline
\multirow{2}{*}{\begin{tabular}[c]{@{}c@{}}RNF\\ -S13\end{tabular}} & 69.8 & 68.8 & 70.9 & 69.7 & 69.9 & 70.0 & 70.7 & 67.1 & 72.4 & 71.7 & 69.1 & 66.6 & 61.7 & 57.6 & \multirow{2}{*}{10.8} \\
 & 66.3 & 65.5 & 67.1 & 66.3 & 66.2 & 65.1 & 66.5 & 63.8 & 69.6 & 68.5 & 64.2 & 63.8 & 58.5 & 51.4 &  \\ \hline
\multirow{2}{*}{\begin{tabular}[c]{@{}c@{}}RNF\\ -C13\end{tabular}} & 77.6 & 76.8 & 78.5 & 76.7 & 78.6 & 78.0 & 79.3 & 73.9 & 78.1 & 79.2 & 76.8 & 75.5 & 70.5 & 67.6 & \multirow{2}{*}{10.5} \\
 & 73.9 & 73.4 & 74.5 & 73.0 & 75.0 & 74.0 & 75.9 & 70.2 & 75.1 & 75.8 & 71.7 & 72.3 & 66.9 & 61.9. &  \\ \hline
\multirow{2}{*}{\begin{tabular}[c]{@{}c@{}}RNF\\ -D13\end{tabular}} & 69.9 & 68.7 & 71.3 & 69.5 & 70.4 & 69.8 & 70.9 & 67.6 & 72.7 & 71.8 & 69.1 & 66.6 & 62.7 & 59.2 & \multirow{2}{*}{10.8} \\
 & 65.4 & 64.5 & 66.3 & 65.3 & 65.5 & 79.2 & 80.3 & 63.4 & 69.4 & 67.3 & 63.5 & 63.2 & 58.8 & 52.8 &  \\ \hline
\multirow{2}{*}{\begin{tabular}[c]{@{}c@{}}RNF\\ -S23\end{tabular}} & 69.9 & 68.7 & 71.4 & 69.6 & 70.3 & 70.0 & 70.7 & 67.2 & 72.1 & 71.9 & 69.1 & 66.7 & 62.0 & 58.7 & \multirow{2}{*}{10.3} \\
 & 66.4 & 65.6 & 67.3 & 66.0 & 66.9 & 66.0 & 67.8 & 63.0 & 69.1 & 68.5 & 64.3 & 63.6 & 60.0 & 54.6 &  \\ \hline
\multirow{2}{*}{\begin{tabular}[c]{@{}c@{}}{RNF}\\ -D23\end{tabular}} & 70.1 & 69.7 & 72.7 & 70.4 & 70.6 & 70.1 & 71.2 & 67.1 & 72.9 & 72.3 & 70.4 & 66.9 & 62.5 & 59.3 & \multirow{2}{*}{10.2} \\
 & 66.1 & 65.1 & 67.2 & 65.4 & 67.3 & 66.1 & 67.3 & 63.4 & 70.4 & 68.9 & 64.2 & 63.3 & 59.2 & 52.6 &  \\ \hlineB{3}
\end{tabular}
\caption{\label{tab:e_3}Proj41-based LLDN performance for backbones with various depth.}
}
\end{table*}

\begin{table*}[htb!]
{
\tabcounterint
\small
\centering
\begin{tabular}{c|c|c|c|c|c|c|c|c|c|c|c|c|c|c|c}
\hlineB{3}
{\begin{tabular}[c]{@{}c@{}}Back-\\ bone\end{tabular}} & {Total} & {Day} & {Night} & {\begin{tabular}[c]{@{}c@{}}Ur-\\ ban\end{tabular}} & {\begin{tabular}[c]{@{}c@{}}High-\\ way\end{tabular}} & {\begin{tabular}[c]{@{}c@{}}Nor-\\ mal\end{tabular}} & {\begin{tabular}[c]{@{}c@{}}Gentle \\ Curve\end{tabular}} & {\begin{tabular}[c]{@{}c@{}}Sharp \\ Curve\end{tabular}} & {\begin{tabular}[c]{@{}c@{}}Mer-\\ ging\end{tabular}} & {\begin{tabular}[c]{@{}c@{}}No\\ Occ\end{tabular}} & {\begin{tabular}[c]{@{}c@{}}Occ\\ 1\end{tabular}} & {\begin{tabular}[c]{@{}c@{}}Occ\\ 2\end{tabular}} & {\begin{tabular}[c]{@{}c@{}}Occ\\ 3\end{tabular}} & {\begin{tabular}[c]{@{}c@{}}Occ\\ 4-6\end{tabular}} & {FPS} \\ \hlineB{2}
\multirow{2}{*}{\begin{tabular}[c]{@{}c@{}}GFC\\ -T1\end{tabular}} & 64.3 & 63.7 & 64.9 & 61.0 & 68.2 & 65.3 & 66.2 & 51.4 & 64.6 & 64.9 & 63.2 & 65.6 & 59.8 & 56.1 & \multirow{2}{*}{14.0}\\
 & 62.4 & 61.9 & 62.9 & 58.9 & 68.2 & 63.5 & 64.5 & 48.5 & 63.3 & 63.1 & 60.3 & 64.4 & 58.1 & 54.4 &  \\ \hline
\multirow{2}{*}{\begin{tabular}[c]{@{}c@{}}GFC\\ -T3\end{tabular}} & 76.2 & 76.2 & 76.1 & 73.6 & \textbf{79.2} & 76.8 & 78.7 & 67.0 & 75.0 & 76.7 & 75.5 & 77.1 & 72.2 & 69.2 & \multirow{2}{*}{13.9} \\
 & 74.7 & 75.0 & 74.4 & 72.0 & 78.1 & 75.4& 77.3 & 64.8 & 74.0 & 75.3 & 73.1 & 76.5 & 71.0 & 68.6 &  \\ \hline
\multirow{2}{*}{\begin{tabular}[c]{@{}c@{}}GFC\\ -T5\end{tabular}} & \textbf{78.5} & \textbf{78.5} & \textbf{78.4} & \textbf{77.8} & \textbf{79.2} & \textbf{77.9} & \textbf{79.0} & \textbf{72.5} & \textbf{78.0} & \textbf{79.4} & \textbf{77.8} & \textbf{77.7} & \textbf{74.5} & \textbf{70.2} & \multirow{2}{*}{13.8} \\
 & \textbf{77.3 }& \textbf{77.6} & \textbf{77.0} & \textbf{76.4} & \textbf{78.4} & \textbf{77.6} & \textbf{77.8} & \textbf{70.2} & \textbf{76.9} & \textbf{78.2} & \textbf{76.3} & \textbf{77.2} & \textbf{73.2} & \textbf{69.5} & \\ \hline  
\multirow{2}{*}{\begin{tabular}[c]{@{}c@{}}GFC\\ -M1\end{tabular}} & 64.5 & 63.7 & 65.5 & 59.1 & 71.0 & 65.3 & 69.8 & 50.6 & 59.9 & 65.1 & 63.9 & 66.7 & 60.0 & 50.4 & \multirow{2}{*}{\textbf{16.6}} \\
 & 62.9 & 62.1 & 63.8 & 57.2 & 69.7 & 63.8 & 68.4 & 47.7 & 59.0 & 63.6 & 61.3 & 65.6 & 58.4 & 48.7 & \\ \hline
\multirow{2}{*}{\begin{tabular}[c]{@{}c@{}}GFC\\ -M3\end{tabular}} & 70.0 & 69.8 & 70.1 & 66.0 & 74.7 & 71.0 & 73.7 & 56.7 & 66.4 & 70.6 & 69.4 & 72.0 & 64.6 & 57.5 & \multirow{2}{*}{16.4} \\
 & 60.6 & 61.2 & 59.8 & 56.8 & 65.1 & 62.0 & 63.6 & 44.3 & 57.8 & 61.6 & 58.1 & 63.0 & 56.3 & 42.9 &  \\ \hline
\multirow{2}{*}{\begin{tabular}[c]{@{}c@{}}GFC\\ -M5\end{tabular}} & 74.8 & 74.8 & 74.9 & 72.0 & 78.2 & 75.6& 77.9& 64.6& 72.0 & 75.5 & 74.4 & 76.0 & 69.3 & 65.2 & \multirow{2}{*}{16.3} \\
 & 73.5 & 73.6 & 73.4 & 70.5 & 77.1 & 74.4 & 76.6 & 62.2 & 71.1 & 74.2 & 72.3 & 75.5 & 67.6 & 62.3 & \\ \hline
\multirow{2}{*}{\begin{tabular}[c]{@{}c@{}}{RNF}\\ -C8\end{tabular}} & 70.1 & 69.3 & 71.1 & 67.6 & 73.2 & 69.9 & 75.7 & 62.9 & 70.2 & 71.5 & 70.9 & 69.0 & 61.9 & 50.9 &  \multirow{2}{*}{15.9}\\
 & 22.1 & 22.3 & 21.9 & 25.1 & 18.5 & 23.2 & 19.1 & 17.4 & 23.1 & 25.8 & 16.9 & 16.6 & 12.7 & 8.5 &  \\ \hline
\multirow{2}{*}{\begin{tabular}[c]{@{}c@{}}RNF\\ -S13\end{tabular}} & 64.6 & 62.9 & 66.5 & 59.4 & 70.7 & 51.1 & 72.1 & 51.2 & 63.4 & 65.5 & 65.9 & 65.0 & 56.9 & 44.9 & \multirow{2}{*}{15.7} \\
 & 18.2 & 16.4 & 20.4 & 15.6 & 21.4 & 13.1 & 22.6 & 13.1 & 16.0 & 19.2 & 18.7 & 16.8 & 16.0 & 4.7 & \\ \hline
\multirow{2}{*}{\begin{tabular}[c]{@{}c@{}}RNF\\ -C13\end{tabular}} & 76.8 & 75.9 & 77.8 & 74.5 & 79.6 & 67.6 & 81.4 & 67.6 & 76.6 & 77.9 & 77.9 & 75.7 & 69.5 & 62.5 & \multirow{2}{*}{15.5} \\
 & 40.6 & 39.1 & 42.4 & 40.6 & 40.6 & 32.5 & 43.6 & 32.5 & 42.6 & 43.6 & 38.3 & 35.7 & 32.8 & 20.4 & \\ \hline
\multirow{2}{*}{\begin{tabular}[c]{@{}c@{}}RNF\\ -D13\end{tabular}} & 61.4 & 60.0 & 62.9 & 56.8 & 66.8 & 61.0 & 69.5 & 51.8 & 60.0 & 62.3 & 62.7 & 61.2 & 53.8 & 42.6 & \multirow{2}{*}{15.4} \\
 & 16.5 & 14.8 & 18.4 & 13.5 & 20.0 & 15.8 & 22.6 & 12.2 & 13.9 & 17.4 & 16.7 & 14.5 & 14.2 & 4.5 &  \\ \hline
\multirow{2}{*}{\begin{tabular}[c]{@{}c@{}}RNF\\ -S23\end{tabular}} & 58.5 & 56.9 & 60.4 & 52.6 & 65.6 & 59.2 & 64.9 & 42.2 & 55.8 & 59.1 & 59.1 & 59.9 & 51.5 & 43.6 &  \multirow{2}{*}{15.3}\\
 & 31.9 & 29.9 & 34.2 & 28.4 & 36.1 & 32.0 & 42.2 & 23.2 & 30.2 & 32.6 & 32.1 & 32.0 & 28.5 & 17.3 &  \\ \hline
\multirow{2}{*}{\begin{tabular}[c]{@{}c@{}}RNF\\ -D23\end{tabular}} & 63.2 & 61.6 & 65.0 & 58.6 & 68.6 & 51.1 & 70.7 & 51.3 & 62.2 & 64.2 & 64.3 & 63.1 & 55.1 & 43.9 & \multirow{2}{*}{15.2} \\
 & 19.0 & 17.0 & 21.2 & 17.2 & 21.1 & 13.5 & 23.7 & 13.5 & 17.2 & 20.2 & 20.2 & 19.1 & 16.7 & 4.9 & \\ \hlineB{3}
\end{tabular}
\caption{\label{tab:e_4}Pillars-based LLDN performance for backbones with various depth.}
}
\end{table*}

\begin{table*}[htb]
{
\tabcounterint
\small
\centering
\begin{tabular}{c|c|c|c|c|c|c|c|c|c|c|c|c|c|c|c}
\hlineB{3}
{\begin{tabular}[c]{@{}c@{}}Back-\\ bone\end{tabular}} & {Total} & {Day} & {Night} & {\begin{tabular}[c]{@{}c@{}}Ur-\\ ban\end{tabular}} & {\begin{tabular}[c]{@{}c@{}}High-\\      way\end{tabular}} & {\begin{tabular}[c]{@{}c@{}}Nor-\\ mal\end{tabular}} & {\begin{tabular}[c]{@{}c@{}}Gen.\\Cur.\end{tabular}} & {\begin{tabular}[c]{@{}c@{}}Sha.\\Cur.\end{tabular}} & {\begin{tabular}[c]{@{}c@{}}Mer-\\ ging\end{tabular}} & {\begin{tabular}[c]{@{}c@{}}No\\ Occ\end{tabular}} & {\begin{tabular}[c]{@{}c@{}}Occ\\ 1\end{tabular}} & {\begin{tabular}[c]{@{}c@{}}Occ\\ 2\end{tabular}} & {\begin{tabular}[c]{@{}c@{}}Occ\\ 3\end{tabular}} & {\begin{tabular}[c]{@{}c@{}}Occ\\ 4-6\end{tabular}} & {FPS} \\ \hlineB{2}
\multirow{2}{*}{\begin{tabular}[c]{@{}c@{}}$P$8\\ $N_h$512\end{tabular}} & \textbf{82.1} & \textbf{82.2} & \textbf{82.0} & \textbf{81.7} & \textbf{82.5} & \textbf{82.5} & 82.2 & \textbf{78.0} & \textbf{81.0} & \textbf{82.9} & \textbf{81.4} & \textbf{82.3} & \textbf{78.7} & \textbf{75.9} & \multirow{2}{*}{11.6} \\
 & \textbf{81.1} & \textbf{81.4} & \textbf{80.7} & \textbf{80.6} & \textbf{81.7} & \textbf{81.5} & \textbf{83.0} & \textbf{76.7} & \textbf{80.1} & \textbf{81.9} & \textbf{81.4} & \textbf{81.3} & \textbf{78.7} & \textbf{75.5} &  \\ \hline
\multirow{2}{*}{\begin{tabular}[c]{@{}c@{}}$P$16\\ $N_h$512\end{tabular}} & 80.2 & 79.3 & 81.2 & 78.4 & 82.4 & 81.0 & 82.4 & 73.3 & 78.9 & 80.7 & 80.5 & 79.7 & 76.6 & 75.6 & \multirow{2}{*}{\textbf{11.9}} \\
 & 78.1 & 77.6 & 78.7 & 75.8 & 80.9 & 79.0 & 80.5 & 70.4 & 76.9 & 78.5 & 77.7 & 78.5 & 75.1 & 74.6 &  \\ \hline
\multirow{2}{*}{\begin{tabular}[c]{@{}c@{}}$P$8\\ $N_h$128\end{tabular}} & 81.5 & 81.5 & 81.5 & 81.1 & 82.0 & 81.9 & \textbf{82.8} & 76.9 & 80.2 & 82.5 & 81.2 & 80.6 & 76.0 & 74.3 & \multirow{2}{*}{11.8} \\
 & 75.3 & 75.7 & 74.9 & 74.5 & 76.4 & 76.2 & 76.6 & 66.2 & 74.6 & 76.3 & 73.9 & 75.5 & 70.3 & 68 &\\ \hline
\multirow{2}{*}{\begin{tabular}[c]{@{}c@{}}$P$8\\ $N_h$2048\end{tabular}} & 76.6 & 75.9 & 77.4 & 75.5 & 77.8 & 77.4 & 78.3 & 67.3 & 76.1 & 77.8 & 76.7 & 74.9 & 71.3 & 64.7 & \multirow{2}{*}{10.7} \\
 & 61.2 & 60.4 & 62.1 & 59.6 & 63.1 & 62.3 & 61.9 & 50.2 & 62.6 & 63.1 & 58.5 & 60.1 & 55.5 & 44.9 &\\ \hlineB{3}
\end{tabular}
\caption{\label{tab:e_5}Performance of Proj28-GFC-T3 for various hidden dimension and patch sizes, where $P$ and $N_h$ represent the patch size and the hidden dimension size with default value 8 and 512, respectively.}
}
\end{table*}

\begin{table*}[htb!]
{
\tabcounterint
\small
\centering
\begin{tabular}{c|c|c|c|c|c|c|c|c|c|c|c|c|c|c|c}
\hlineB{3}
{\begin{tabular}[c]{@{}c@{}}Back-\\ bone\end{tabular}} & {Total} & {Day} & {Night} & {\begin{tabular}[c]{@{}c@{}}Ur-\\ ban\end{tabular}} & {\begin{tabular}[c]{@{}c@{}}High-\\      way\end{tabular}} & {\begin{tabular}[c]{@{}c@{}}Nor-\\ mal\end{tabular}} & {\begin{tabular}[c]{@{}c@{}}Gen.\\Cur.\end{tabular}} & {\begin{tabular}[c]{@{}c@{}}Sha.\\Cur.\end{tabular}} & {\begin{tabular}[c]{@{}c@{}}Mer-\\ ging\end{tabular}} & {\begin{tabular}[c]{@{}c@{}}No\\ Occ\end{tabular}} & {\begin{tabular}[c]{@{}c@{}}Occ\\ 1\end{tabular}} & {\begin{tabular}[c]{@{}c@{}}Occ\\ 2\end{tabular}} & {\begin{tabular}[c]{@{}c@{}}Occ\\ 3\end{tabular}} & {\begin{tabular}[c]{@{}c@{}}Occ\\ 4-6\end{tabular}} & {FPS} \\ \hlineB{2}
\multirow{2}{*}{\begin{tabular}[c]{@{}c@{}}$P$8\\ $N_h$512\end{tabular}} & 74.8 & 74.8 & 74.9 & 72.0 & 78.2 & 77.9 & 75.6 & \textbf{64.6} & 72.0 & 75.5 & 74.4 & 76.0 & 69.3 & 65.2 & \multirow{2}{*}{16.3} \\
 & 73.5 & 73.6 & 73.4 & 70.5 & 77.1 & 74.4 & 76.6 & 62.2 & 71.1 & 74.2 & 72.3 & 75.5 & 67.6 & 62.3 & \\ \hline
\multirow{2}{*}{\begin{tabular}[c]{@{}c@{}}$P$16\\ $N_h$512\end{tabular}} & 72.2 & 72.0 & 72.4 & 68..8 & 76.2 & 67.0 & 75.9 & 58.4 & 70.2 & 72.6 & 71.5 & 73.4 & 68.9 & 63.4 & \multirow{2}{*}{\textbf{16.6}} \\
 & 65.2 & 65.2 & 65.2 & 61.8 & 69.3 & 73.0 & 67.2 & 49.5 & 64 & 66.2 & 62.8 & 66.8 & 60.5 & 55.8 &  \\ \hline
\multirow{2}{*}{\begin{tabular}[c]{@{}c@{}}$P$8\\ $N_h$128\end{tabular}} & 70.9 & 70.9 & 71.0 & 67.7 & 74.8 & 71.7 & 74.6 & 59.4 & 68.4 & 71.6 & 70.2 & 72.5 & 65.7 & 60.2 & \multirow{2}{*}{16.5} \\
 & 63.1 & 62.7 & 63.4 & 59.4 & 67.4 & 64.1 & 66.0 & 49.2 & 61.7 & 64.1 & 60.7 & 65.4 & 56.7 & 48.5 &  \\ \hline
\multirow{2}{*}{\begin{tabular}[c]{@{}c@{}}$P$8\\ $N_h$2048\end{tabular}} & \textbf{75.6} & \textbf{75.5} & \textbf{75.6} & \textbf{72.9} & \textbf{78.8} & \textbf{76.4} & \textbf{78.8} & 64.5 & \textbf{73.5} & \textbf{76.2} & \textbf{74.8} & \textbf{76.6} & \textbf{71.2} & \textbf{66.5} & \multirow{2}{*}{14.7} \\
 & \textbf{74.1} & \textbf{74.4} & \textbf{73.8} & \textbf{71.2} & \textbf{77.6} & \textbf{75.0} & \textbf{77.2} & \textbf{62.6} & \textbf{72.6} & \textbf{74.8} & \textbf{72.5} & \textbf{76.1} & \textbf{69.7} & \textbf{68.2} &  \\ \hlineB{3}
\end{tabular}
\caption{\label{tab:e_6}Performance of Pillars-GFC-M5 for various hidden dimension and patch sizes; where $P$ and $N_h$ represent the patch size and the hidden dimension size with default value 8 and 512, respectively.}
}
\end{table*}

\noindent \textbf{Ablations on Hidden Dimension.} As shown in the Table \ref{tab:e_5} and \ref{tab:e_6}, we perform ablation studies on different hidden dimension size for Proj28-GFC-T3 and Pillars-GFC-M5, which are the best performing model of the proposed LLDN-GFC and its low computational alternative, respectively. As denoted in Section B.1, the hidden dimension $N_h$ is the number of channels for each patch after the per-patch linear transform, indicating that higher value of hidden dimension leads to higher model capacity per each grid.

Table \ref{tab:e_5} shows the performance for various hidden dimension $N_h$ of Proj28-GFC-T3; $N_h$=512 outperforms other variants, such as $N_h$=128 and $N_h$=2048. On the other hand, since Pillars-GFC-M5 requires more model capacity per each grid than Proj-GFC-T, Pillars-GFC-M with $N_h$=2048 outperforms that with $N_h$=512.

\noindent \textbf{Ablations on Patch Size.} We also perform ablation studies on the patch size of the Proj28-GFC-T3 and Pillars-GFC-M5. The results in Table \ref{tab:e_5} and \ref{tab:e_6} show that there is a significant performance drop as P is increased from 8 to 16. This is because when the patch size is increased to 16 (from 8), the number of grids covered by a patch increases four times, so that the GFC has to extract global features from a map with four times lower resolution.

\vspace{5mm}
{\large\noindent\textbf{D. Qualitative Results Visualization}}
\vspace{2mm}

In addition to the numerical results in Section 4.1, we provide qualitative results of the proposed LLDN-GFC, Proj28-GFC-T3, its low computational alternative, Pillars-GFC-M5, and the conventional CNN-based LLDNs, such as Proj28-RNF-S13, Proj28-RNF-C13, and Pillars-RNF-C13, using visualization.

\vspace{2mm}
{\large\noindent\textbf{D.1. Qualitative Results}}
\vspace{2mm}

\begin{figure}
\figcounterint
    \centering
    \vspace{3mm}
    \includegraphics[width=1.0\columnwidth]{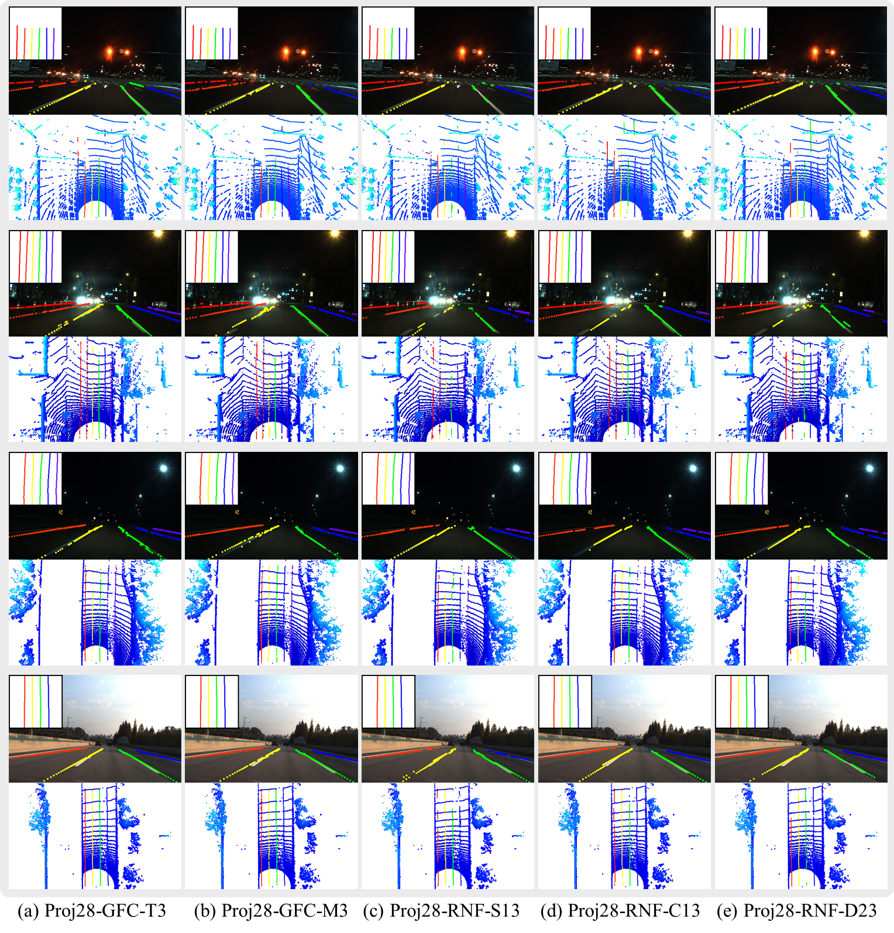}
    \caption{Lane detection performance comparison for LLDNs with Proj28 for images with moderate difficulty (e.g., normal, no occlusion, and gentle curve).}
    \label{fig:f_1}
\end{figure}

\begin{figure}
\figcounterint
    \centering
    \vspace{4mm}
    \includegraphics[width=1.0\columnwidth]{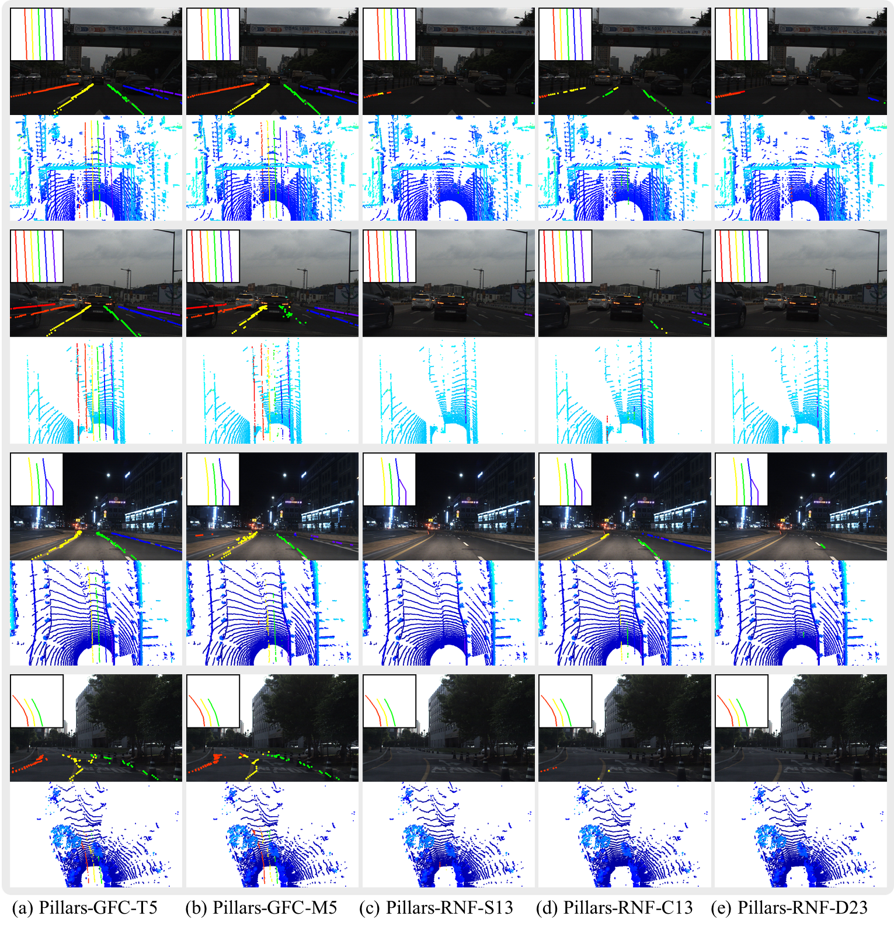}
    \caption{Lane detection performance comparison for LLDNs with Pillars for images with high difficulty (e.g., occlusion, merging, and sharp curve).}
    \label{fig:f_4}
\end{figure}

Fig. \ref{fig:f_1}$\sim$\ref{fig:f_4} in this subsection has 4 rows and 5 columns, where each row shows inference results for different scenes (conditions) and each column shows inference results for different GFCs. Each inference result has upper and lower plots for the projection of inference results into the front view image with true labels in the upper left corner and for the inference on top of the BEV point cloud, respectively.

\begin{figure}
\figcounterint
    \centering
    \vspace{4mm}
    \includegraphics[width=1.0\columnwidth]{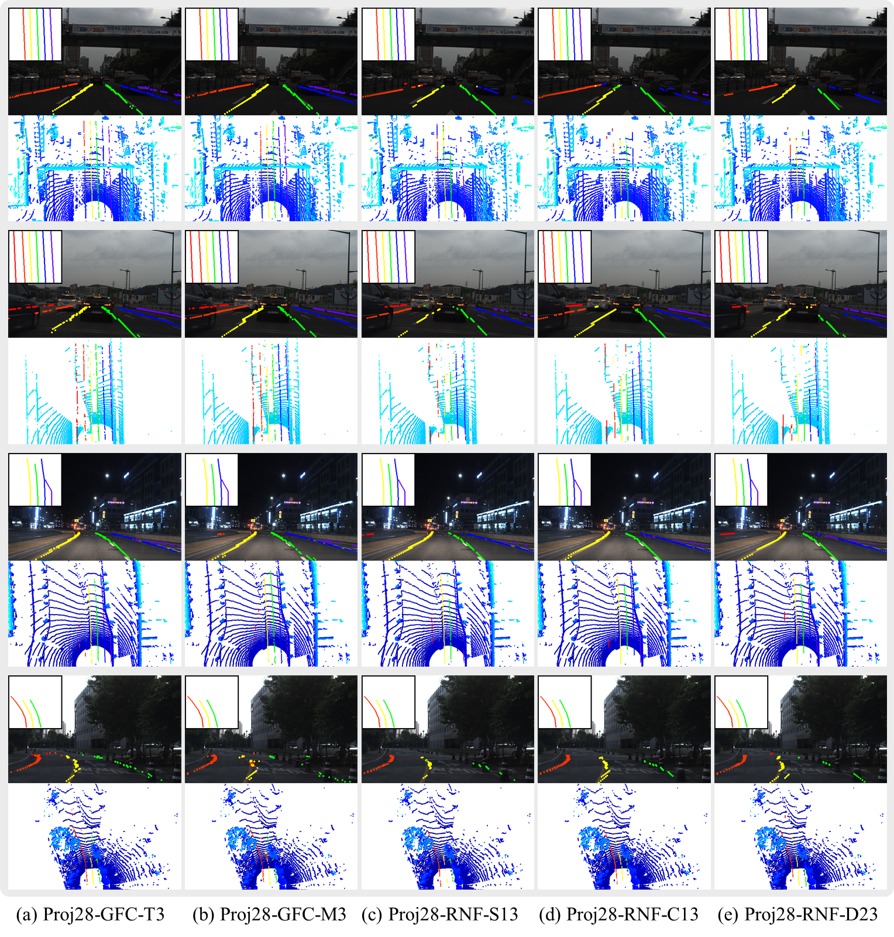}
    \caption{Lane detection performance comparison for LLDNs with Proj28 for images with high difficulty (e.g., occlusion, merging, and sharp curve).}
    \label{fig:f_2}
\end{figure}

\begin{figure}
\figcounterint
    \centering
    \vspace{3mm}
    \includegraphics[width=1.0\columnwidth]{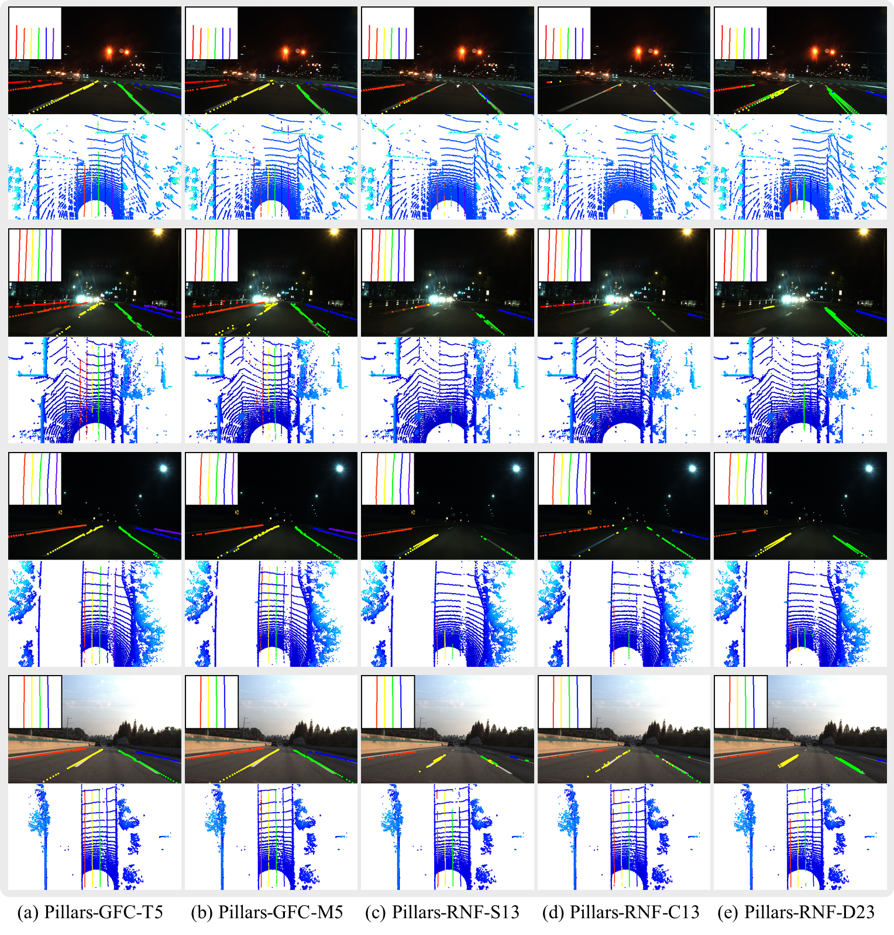}
    \caption{Lane detection performance comparison for LLDNs with Pillars for images with moderate difficulty (e.g., normal, no occlusion, and gentle curve).}
    \label{fig:f_3}
\end{figure}

Fig. \ref{fig:f_1} and \ref{fig:f_2} show inference results of LLDNs with Proj28 for scenes with moderate (e.g., normal, no occlusion, and gentle curve) and severe (e.g., occlusion, merging, and sharp curve) lane detection difficulties, respectively, while Fig. \ref{fig:f_3} and \ref{fig:f_4} show inference results of LLDNs with Pillars.

In all figures shown in this subsection, LLDNs based on GFC-T and GFC-M (shown in (a) and (b) of figures, respectively) show better performance than other LLDNs (in (c), (d), and (e) of figures) regardless of the lane detection difficulties and BEV encoders (i.e., Proj28 and Pillars). For example, plots in (a) and (b) show a strong lane detection performance even for images of severe occlusion, where a good portion of point cloud data are missing.

\vspace{2mm}
{\large\noindent\textbf{D.2. Qualitative Heatmaps}}
\vspace{2mm}

\begin{figure}[tb!]
    \figcounterint
    \centering
    \includegraphics[width=1.0\columnwidth]{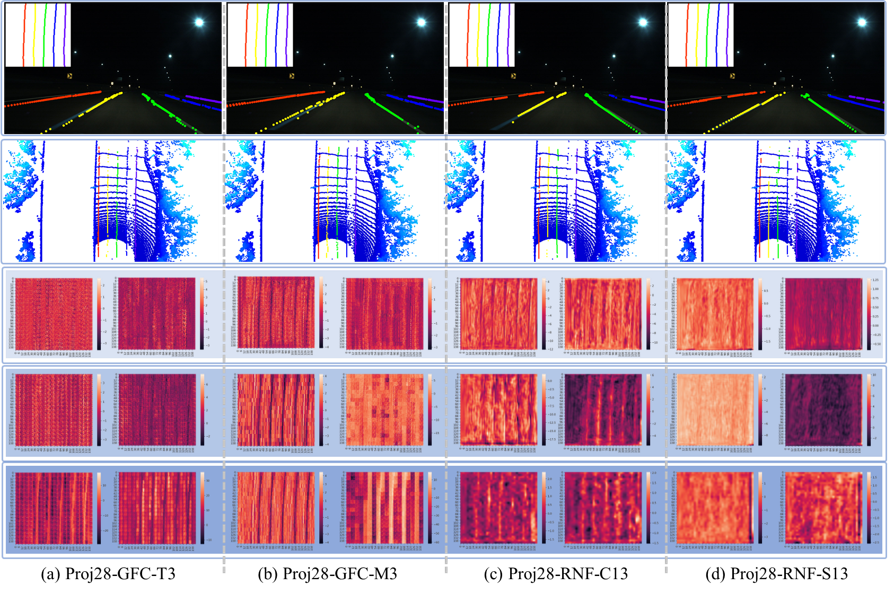}
    \caption{Comparison of the lane detection performance of the proposed LLDN-GFC, Proj28-GFC-T3, to the Proj28-GFC-T3 and other CNN-based LLDNs (Proj28-RNF-C13 and Proj28-RNF-S13) for curved lanes.}
\label{fig:g_1}
\end{figure}

\begin{figure}[tb!]
    \figcounterint
    \centering
    \includegraphics[width=1.0\columnwidth]{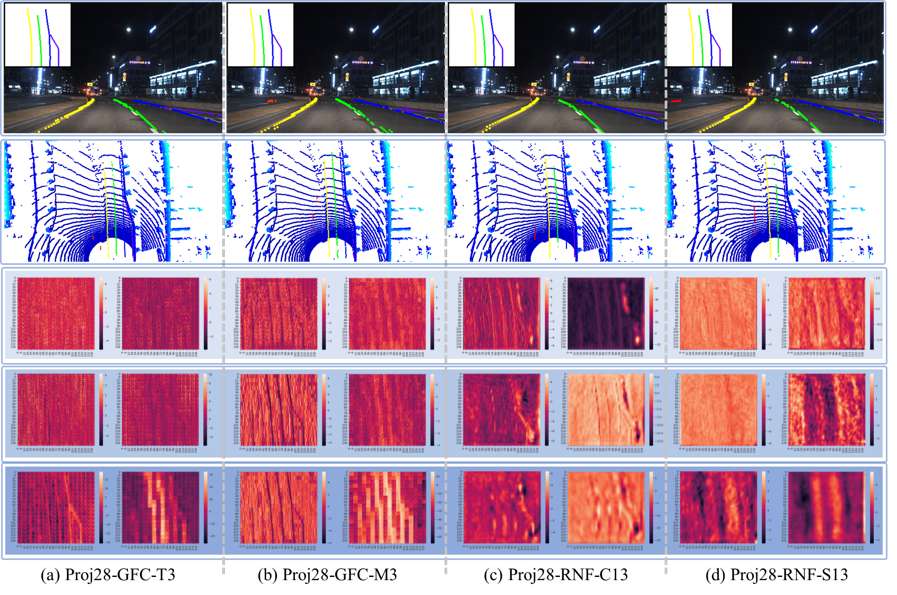}
    \caption{Comparison of the lane detection performance of the proposed LLDN-GFC, Proj28-GFC-T3, to the Proj28-GFC-M3 and other CNN-based LLDNs (Proj28-RNF-C13 and Proj28-RNF-S13) for merging lanes.}
\label{fig:g_2}
\end{figure}

\begin{figure}[tb!]
    \figcounterint
    \centering
    \includegraphics[width=1.0\columnwidth]{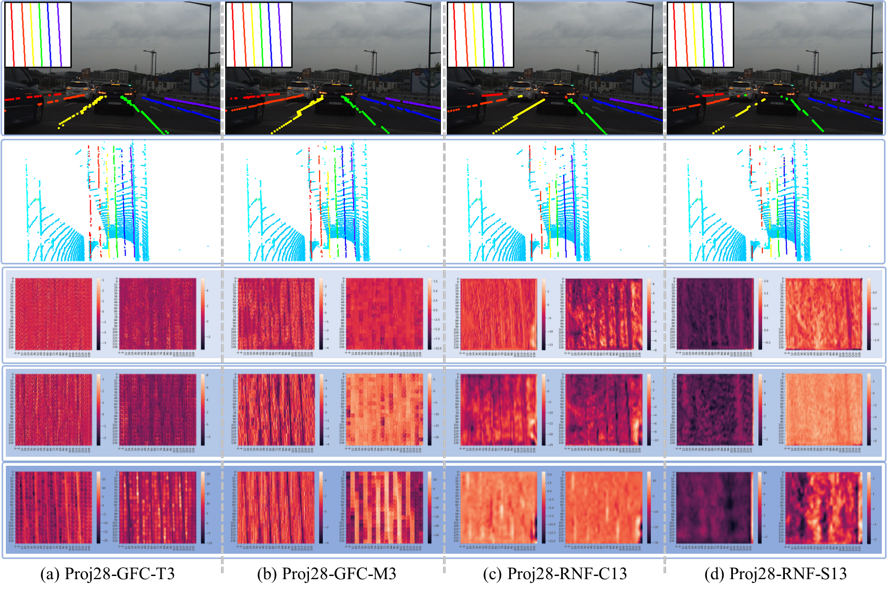}
    \caption{Comparison of the lane detection performance of the proposed LLDN-GFC, Proj28-GFC-T3, to the Proj28-GFC-M3 and other CNN-based LLDNs (Proj28-RNF-C13 and Proj28-RNF-S13) for occluded lanes.}
\label{fig:g_3}
\end{figure}

We emphasize the performance of the proposed GFC, GFC-T, and its low computational alternative, GFC-M, using the visualization of the heatmaps for occlusion scenes as shown in Fig. \ref{heatmap} in Section 4.1. In addition to the results in Fig. \ref{heatmap}, we provide more visualization of heatmaps for various difficult scenes, such as curved lanes, merging lanes, and other severe occlusion cases, to emphasize the superior performance of the proposed GFC and its low-computational alternative GFC.

All of the figures in this subsection follow the same format used for Fig. \ref{heatmap} in Section 4.1. The four columns are inference results for (a) Proj28-GFC-T3, (b) Proj28-GFC-M3, (c) Proj28-RNF-C13, and (d) Proj28-RNF-S13. The first row shows the projection of inference results into the front view image with true labels in the upper left corner, and the second row shows the inference on top of the BEV point cloud. From the 3rd to 5th row, we show the heatmap of the 1st, 2nd, and 3rd block output feature map of the GFC (e.g., 1st, 2nd, and 3rd Transformer block of Proj28-GFC-T3). Output feature maps at different blocks are resized or reshaped (i.e., in the same way to the function (4-1) in Fig. \ref{gfc_detail}) and two heatmaps are sampled along the channels.

As shown in Fig. \ref{fig:g_1}, Fig. \ref{fig:g_2}, and Fig. \ref{fig:g_3}, both the proposed LLDN-GFC, Proj28-GFC-T3, and the LLDN with the low computational alternative GFC, GFC-M3, demonstrate three advantages described in Section 4.1 for curved lanes, merging lanes, and other occluded lanes. The three advantages are (1) better resolution as the network deepens, (2) distinctive color difference between lane and non-lane positions, and (3) predicting the shape of the lane even in presence of occlusion.

\vspace{5mm}
{\large\noindent\textbf{E. LLDN vs Heuristic Method}}
\vspace{2mm}

\begin{figure}[b!]
{
\figcounterint
\centering
\includegraphics[width=1.0\columnwidth]{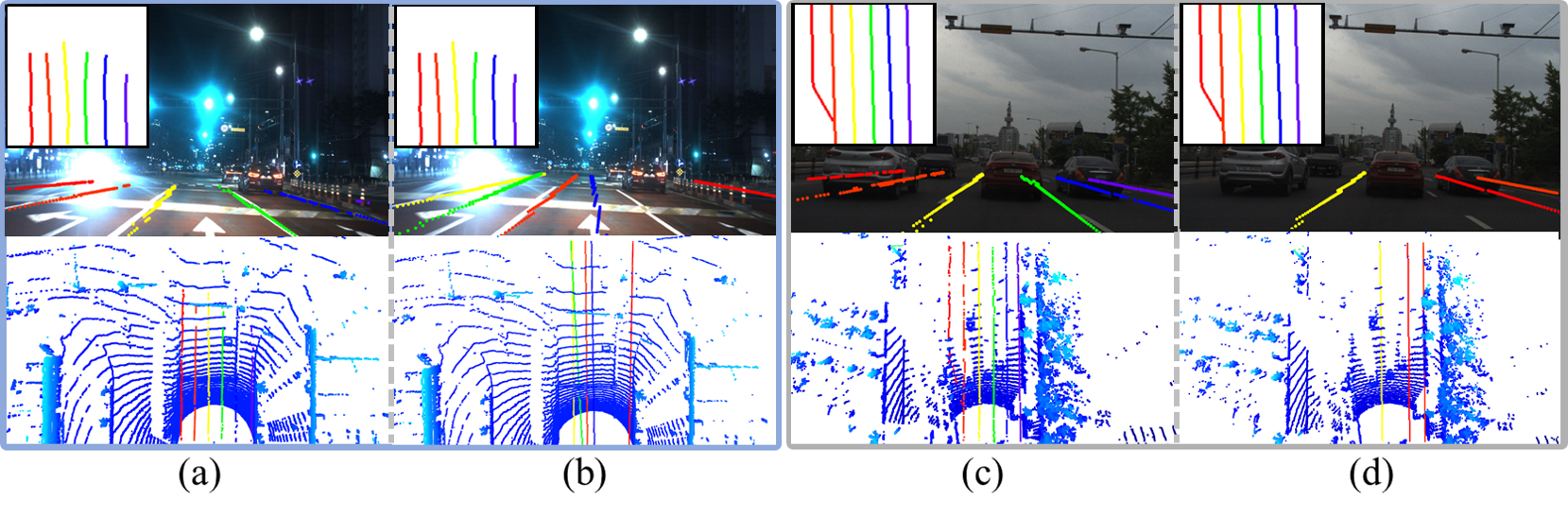}
\caption{Comparison between the proposed LLDN-GFC (Proj28-GFC-T3) (in (a) and (c)) and heuristic Lidar lane detection (in (b) and (d)). When a strong source of illumination appears on the scene, (b) the heuristic method fails, but (a) the proposed LLDN-GFC is not affected. When lane marks are occluded, (d) the heuristic method cannot infer the lanes, but (c) the proposed LLDN-GFC is able to infer the occluded lane lines.}
\label{heuristic}
}
\end{figure}

In the heuristic Lidar lane detection techniques, we first project pointcloud into a BEV image and apply thresholding to remove low-intensity points \cite{heu_lidar_3}. The remaining points are then clustered using, for example, DBSCAN \cite{dbscan} and then fitted by the first order polynomial to create smooth lane lines.

In the experiments, we observe multiple instances when the heuristic technique is unreliable; First, when a strong light illuminates a spot on the ground, as shown in Fig. \ref{heuristic} (b), it results in false positives (FPs). Second, when lane marks are occluded, the heuristic Lidar lane detection cannot infer the presence of lane marks, leading to a high false negatives (FNs), as shown in Fig. \ref{heuristic} (d). However, the proposed LLDN-GFC can produce reliable lane detection results for the two scenarios. As the LLDN-GFC learns global context features of the scene, a bright illuminated road spot or partial occlusion of lane lines hardly degrade the lane detection performance.

{\small
\bibliographystyle{ieee_fullname}
\bibliography{main}
}

\end{document}